\documentclass[10pt,twocolumn,letterpaper]{article}

\usepackage{cvpr}              % To produce the CAMERA-READY version
\usepackage[accsupp]{axessibility}  % Improves PDF readability for those with disabilities.
\usepackage{graphicx}
\usepackage{amsmath}
\usepackage{amssymb}
\usepackage{booktabs}
\usepackage[colorlinks]{hyperref}
\usepackage[capitalize]{cleveref}
\usepackage{overpic}
\usepackage[ruled,linesnumbered]{algorithm2e}

\usepackage{times}
\usepackage{epsfig}
\usepackage{graphicx}
\usepackage{amsmath}
\usepackage{mathrsfs}
\usepackage{amssymb}
\usepackage{multirow}
\usepackage{graphics}
\usepackage{threeparttable}
\usepackage{color}
\usepackage[normalem]{ulem}
\usepackage{float}
\usepackage{amsfonts}
\usepackage{bm}
\usepackage{bbm}
\usepackage{enumitem}
\usepackage[numbers]{natbib}
\usepackage{subcaption}
\usepackage{array}
\usepackage{colortbl}
\usepackage{pifont}

\usepackage{mathtools}
\usepackage{siunitx}
\usepackage[nopar]{lipsum}
\usepackage{listings}
\usepackage[export]{adjustbox}
\usepackage{verbatim}
\usepackage{colortbl}
\usepackage[accsupp]{axessibility}  % Improves PDF readability for those with disabilities.

\usepackage{etoolbox}
\makeatletter
\let\@algcomment\relax
\newcommand\algcomment[1]{\def\@algcomment{\footnotesize#1}}
\renewcommand\fs@ruled{\def\@fs@cfont{\bfseries}\let\@fs@capt\floatc@ruled
	\def\@fs@pre{\hrule height.8pt depth0pt \kern2pt}%
	\def\@fs@post{}%
	\def\@fs@mid{\kern2pt\hrule\kern2pt}%
	\let\@fs@iftopcapt\iftrue}
\makeatother
\usepackage[dvipsnames]{xcolor}
\newcommand{\stdvu}[1]{\scriptsize{\color{darkgray}(#1)} {\color{ForestGreen}$\uparrow$}}
\newcommand{\stdvd}[1]{\scriptsize{\color{darkgray}(#1)} {\color{red}$\downarrow$}}
\newcommand{\stdvw}[1]{\scriptsize{\color{darkgray}(#1)} {\color{white}$\downarrow$}}
\definecolor{mypink}{rgb}{.99,.91,.95}
\usepackage{tabulary}
\newcolumntype{I}{!{\vrule width 1pt}}

\newcolumntype{x}[1]{>{\centering\arraybackslash}p{#1pt}}
\newcolumntype{y}[1]{>{\raggedright\arraybackslash}p{#1pt}}
\newcolumntype{z}[1]{>{\raggedleft\arraybackslash}p{#1pt}}
\newlength\savewidth

\newcommand{\myhyperlink}[3][black]{\hyperlink{#2}{\color{#1}{#3}}}
\usepackage[capitalize]{cleveref}
\crefname{section}{§}{§§}
\Crefname{section}{§}{§§}

\usepackage[capitalize]{cleveref}
\crefname{section}{Sec.}{Secs.}
\Crefname{section}{Section}{Sections}
\Crefname{table}{Table}{Tables}
\crefname{table}{Tab.}{Tabs.}

\def\cvprPaperID{2598} % *** Enter the CVPR Paper ID here
\def\confName{CVPR}
\def\confYear{2023}

\begin{document}

%%%%%%%%% TITLE - PLEASE UPDATE
\title{Learning Federated Visual Prompt in Null Space for MRI Reconstruction}

\author{Chun-Mei Feng$^1$\quad Bangjun Li$^2$\quad Xinxing Xu$^{1,*}$\quad Yong Liu$^{1,}$\thanks{Corresponding author.}\quad Huazhu Fu$^1$\quad Wangmeng Zuo$^3$ \\
$^1$Institute of High Performance Computing (IHPC), \\Agency for Science, Technology and Research (A*STAR), Singapore\\
$^2$School of Information Science and Engineering, Shangdong University, China\\
$^3$Harbin Institute of Technology, Harbin, China\\
% Institution1 address\\
{\tt\small strawberry.feng0304@gmail.com, \{xuxinx,liuyong\}@ihpc.a-star.edu.sg}
\\ {\small {\url{https://github.com/chunmeifeng/FedPR}}
}
% {\tt\small xuxinx@ihpc.a-star.edu.sg}
% {\tt\small liuyong@ihpc.a-star.edu.sg}
%
%
%
% For a paper whose authors are all at the same institution,
% omit the following lines up until the closing ``}''.
% Additional authors and addresses can be added with ``\and'',
% just like the second author.
% To save space, use either the email address or home page, not both
% \and
% Bangjun Li\\
% Institution2\\
% First line of institution2 address\\
% {\tt\small secondauthor@i2.org}
}
\maketitle

% Chun-Mei Feng, Institute of High Performance Computing (IHPC), Agency for Science, Technology and Research (A*STAR), 1 Fusionopolis Way, #16-16 Connexis, Singapore 138632, Republic of Singapore, strawberry.feng0304@gmail.com

% Bangjun Li, School of Information Science and Engineering, Shangdong University   libangjun@mail.sdu.edu.cn

% Xinxing Xu, Institute of High Performance Computing (IHPC), Agency for Science, Technology and Research (A*STAR), 1 Fusionopolis Way, #16-16 Connexis, Singapore 138632, Republic of Singapore, xuxinx@ihpc.a-star.edu.sg

% Yong Liu, Institute of High Performance Computing (IHPC), Agency for Science, Technology and Research (A*STAR), 1 Fusionopolis Way, #16-16 Connexis, Singapore 138632, Republic of Singapore, liuyong@ihpc.a-star.edu.sg

% Wangmeng Zuo, Harbin Institute of Technology, China, wmzuo@hit.edu.cn

% Huazhu Fu, Institute of High Performance Computing (IHPC), Agency for Science, Technology and Research (A*STAR), 1 Fusionopolis Way, #16-16 Connexis, Singapore 138632, Republic of Singapore, hzfu@ieee.com

%%%%%%%%% ABSTRACT
\begin{abstract}

Federated Magnetic Resonance Imaging (MRI) reconstruction enables multiple hospitals to collaborate distributedly without aggregating local data, thereby protecting patient privacy.
However, the data heterogeneity caused by different MRI protocols, insufficient local training data, and limited communication bandwidth inevitably impair global model convergence and updating.
% Typically, tackling down the catastrophic forgetting can accelerate the convergence of FL because overfitting local data distribution always results in an inaccurate global model.
% Recently, prompt tuning has emerged as a new paradigm that achieves excellent performance by only training small-scale soft prompts in frozen pre-trained models.
% We reasoned that combining prompt with a powerful pre-trained model could modify the paradigm of FL to significantly reduce communication costs while achieving competitive performance on a limited number of local data.
In this paper, we propose a new algorithm, FedPR, to learn federated visual prompts in the null space of global prompt for MRI reconstruction.
FedPR is a new federated paradigm that adopts a powerful pre-trained model while only learning and communicating the prompts with few learnable parameters, thereby significantly reducing communication costs and achieving competitive performance on limited local data.
%
%only globally shares a small part of local model to 
% FedPR only globally shares a small part of local model and optimizes the local parameters that lie in the null space of global prompts, avoiding forgetting previously acquired knowledge and accelerating convergence.
% Instead of sending back and forth all the model parameters between the server and clients, FedPR only globally shares a small part of the local model, which significantly reduces communication costs.
Moreover, to deal with catastrophic forgetting caused by data heterogeneity, 
FedPR also updates efficient federated visual prompts that project the local prompts into an approximate null space of the global prompt, thereby suppressing the interference of gradients on the server performance.
%while accelerating convergence.
% uncentered covariance of 
%, avoiding forgetting previously acquired knowledge and accelerating convergence.
Extensive experiments on federated MRI show that FedPR significantly outperforms state-of-the-art FL algorithms with $<6\%$ of communication costs when given the limited amount of local training data. %The code will be released upon acceptance.

% Extensive experiments on federated MRI show that FedPR significantly improves the convergence speed and communication efficiency of FL with only a few local data.

\end{abstract}

%%%%%%%%% BODY TEXT
\vspace{-3pt}
\section{Introduction}\label{sec:intro}
% \vspace{-3pt}
% Although the existing deep learning-based Magnetic Resonance Imaging (MRI) reconstruction technologies have achieved remarkable achievements, it is difficult to collect a large number of pairwise data due to the patient privacy issues and the difficulty of collecting ground-truth images, which limits the development of deep learning-based MRI technology~\cite{guo2021multi}. Fortunately, the advent of federated learning (FL) enables multiple hospitals to train a powerful global model in a distributed manner without sharing private data~\cite{mcmahan2017communication}. In FL, \eg, FedAvg, clients use their local computing power, memory, and private data to train local models independently, while the server aggregates all the local models in each communication round and distributes them to each client again~\cite{mcmahan2017communication}.
Federated Magnetic Resonance Imaging (MRI) reconstruction enables multiple hospitals to train a powerful global model in a distributed manner without sharing private data~\cite{guo2021multi,feng2022specificity,feng2022multi}. In federated MRI, each client (\ie, hospital) uses its local computing power, memory, and private data to train local models independently, while the server aggregates all the local models in each communication round and distributes the global model to each client again~\cite{mcmahan2017communication}.

\begin{figure}[t]
	\vspace{-14pt}
	\begin{center}
		\includegraphics[width=0.8\linewidth]{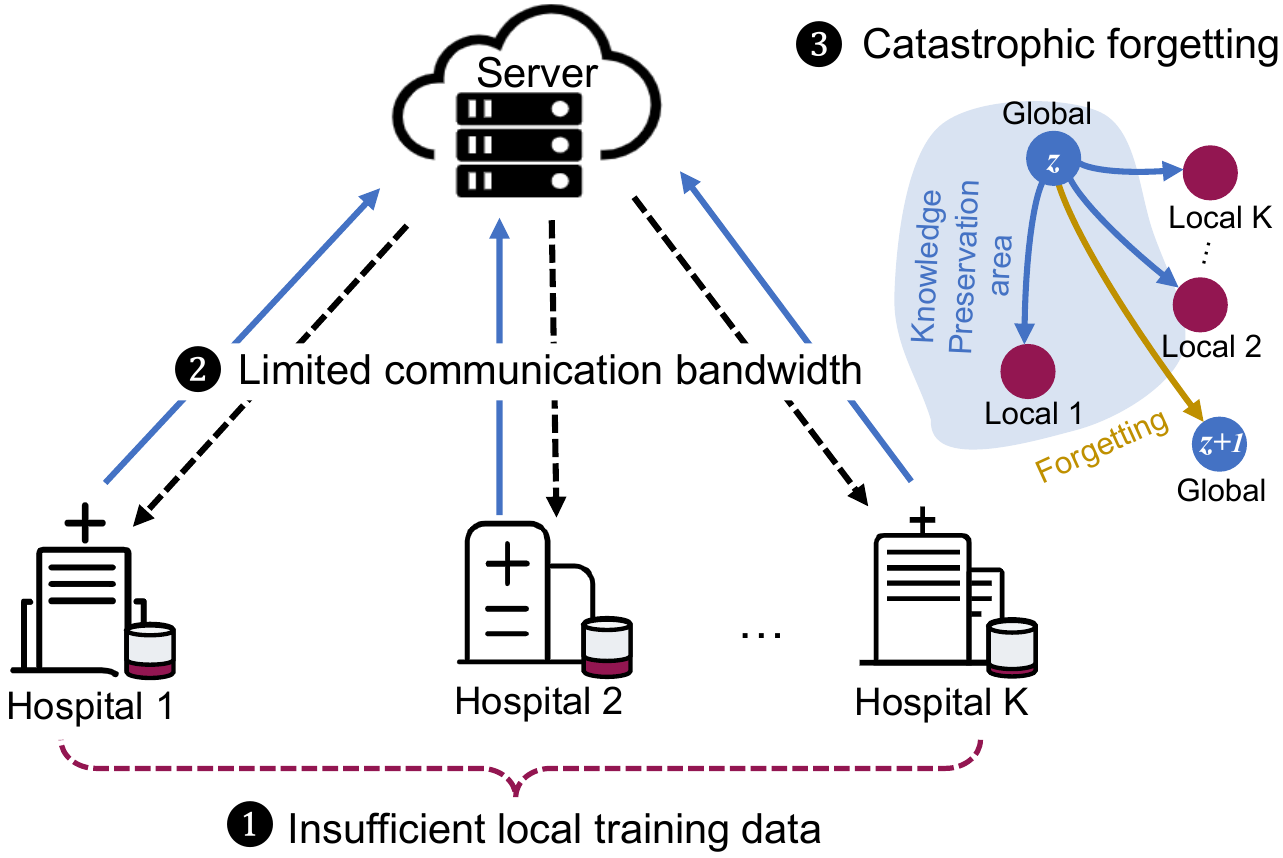}
	\end{center}
	\vspace{-16pt}
	\captionsetup{font=small}
	\caption{\small\textbf{Illustration} of the three \textbf{key issues} in federated MRI.}
	\vspace{-19pt}
	\label{fig:keyissues}
\end{figure}

Existing federated MRI reconstruction techniques usually improve federated learning (FL) by \textit{{enhancing the aggregation process}}~\cite{feng2022specificity} and \textit{{reducing the local parameter's variance}}~\cite{guo2021multi,elmas2022federated}. Such techniques require a large communication bandwidth and sufficient local training data. However, federated MRI often faces two issues,  \hypertarget{Q1}{\ding{182}} \textit{insufficient amount of local training data} due to the difficulty of acquiring the ground-truth of MRI reconstruction~\cite{feng2021exploring,feng2021donet,feng2021dual} and \hypertarget{Q2}{\ding{183}} \textit{limited communication bandwidth} due to unbalanced regional development (see Fig.~\ref{fig:keyissues}). 
%As a result, existing methods cannot systematically address these clinical issues. 
%
To cope with the issue \myhyperlink{Q1}{\ding{182}}, \textit{pre-trained} models have exhibited superior performance, and have shown to \textit{close} the gap between federated and centralized performance~\cite{nguyen2022begin,chen2022pre}.
However, since the model parameters need to be shared between the client and the server for updating, the large number of parameters of pre-trained models will result in a \textit{{huge communication cost}}. 
On contrary, prompt tuning has recently been suggested as a new fine-tuning paradigm by freezing the models and only learning a small number of learnable parameters in the input space~\cite{shin2020autoprompt,schick2020s,jia2022visual}. 
Benefited from pre-trained models, only parameter-efficient prompts are required in learning and communication, and prompt tuning can be conducted with a limited number of samples, making it very appealing in tackling the above two issues (\ie, \myhyperlink{Q1}{\ding{182}} and \myhyperlink{Q2}{\ding{183}}) in federated MRI.
%
%enabling the model to perform well on a limited number of samples. 
%Further, using \textit{pre-trained} models has been shown to \textit{close} the gap between federated and centralized performance~\cite{nguyen2022begin,chen2022pre}. 
%However, since the model parameters need to be shared between the client and the server for updating, the large number of parameters using the pre-training method will result in a \textit{{huge communication cost}}. 
%Inspired by the success of the visual prompt, 
%Thus, we attempt to train federated MRI in this new paradigm to tackle the complex challenges of federated MRI reconstruction, \ie, issues \myhyperlink{Q1}{\ding{182}} and \myhyperlink{Q2}{\ding{183}}.

%In addition to the above two issues, 
Besides, there is another critical issue for federated MRI, \ie, \hypertarget{Q3}{\ding{184}}
%(iii) 
\textit{catastrophic forgetting}, caused by data heterogeneity due to the different imaging protocols of MRI scanners adopted by different clients~\cite{xu2022acceleration,feng2022specificity} (see Fig.~\ref{fig:keyissues}). 
In the local update, the global model from the prior round tends to be overfitted to the local training data, leading to catastrophic forgetting. Analogous to continual learning, several techniques have been presented by introducing proper regularization terms in local models~\cite{shoham2019overcoming,xu2022acceleration,dong2022federated} or retaining previously acquired knowledge by knowledge distillation~\cite{huang2022learn,wei2022knowledge,usmanova2021distillation}. 
However, these strategies require seeking a balance between multiple losses and relying on proxy datasets. 
Instead, Adam-NSC mitigates catastrophic forgetting in continual learning with a new network parameter update rule, \ie, forces the network update to lie in the approximate null space of the input features of previous tasks at each layer~\cite{wang2021training}. However, the null space at feature level is built upon large local training data, which generally cannot be satisfied in federated MRI, \ie, the issue \myhyperlink{Q1}{\ding{182}}. 
Taking the issues \myhyperlink{Q1}{\ding{182}} and \hypertarget{Q3}{\ding{184}} into account, we suggest to optimize local prompts in the approximate null space of global prompts instead of input features, thereby preventing the loss of previously gained global knowledge. 
%enabling the global model to continually update on local data while also

In a nutshell, we explore a new FL paradigm, \ie, FedPR, to learn federated visual prompts for MRI reconstruction. 
% in null space
To begin with, we pre-train the model on a large-scale public dataset.
Given limited amount of local training data, visual prompt tuning is adopted to learn local prompts in a \textit{distributed} manner.
For the issues \myhyperlink{Q1}{\ding{182}} and \myhyperlink{Q2}{\ding{183}}, \textit{\textbf{federated visual prompts}} are introduced to learn a strong global model, where only local and global prompts are learnable and communicated.
%while significantly reducing the training parameters and communication cost.
%
%We use visual prompts to perform downstream tasks only on \textit{\textbf{a few local data}}, where the prompts are trained in a \textit{distributed way}. 
%
As for the issue \myhyperlink{Q3}{\ding{184}}, we  perform singular value decomposition (SVD) on the uncentered covariance matrix of global prompts to obtain the approximate \textit{\textbf{null space}}.
%
%By projecting the local prompts into the null space, %
FedPR tunes only the local parameters in the null space of global prompts, thereby well preserving the \textit{\textbf{prior knowledge}} of previous rounds and resulting in low \textit{\textbf{communication costs}}.
In particular, FedPR achieves a $>4.5$ dB gain in PSNR with less than $6\%$ of communication costs. 
%
%In particular, FedPR achieves a $22.4\%$ improvement in terms of PSNR with less than $4\%$ of trainable parameters. 
To sum up, our contributions are as follows:

\begin{itemize}[leftmargin=*]
	\setlength{\itemsep}{0pt}
	\setlength{\parsep}{-2pt}
	\setlength{\parskip}{-0pt}
	\setlength{\leftmargin}{-15pt}
	\vspace{-6pt}
\item 
%   {We systematically study the key issues in federated MRI reconstruction, \ie, \hypertarget{Q1}{\ding{182}} \textit{data heterogeneity caused by different imaging protocols of MRI scanners}; \hypertarget{Q2}{\ding{183}} \textit{insufficient local training data caused by difficult acquisition of ground-truth data}; and \hypertarget{Q3}{\ding{184}} \textit{insufficient communication bandwidth}, which are the key to catastrophic forgetting and impair global model convergence (see Fig.~\ref{in}.}
    {We propose a \textit{federated visual prompt} algorithm, FedPR, to solve the three key issues in federated MRI. By leveraging powerful pre-trained models and freezing backbone networks in FL, only a small amount of parameters in the input space are trainable, thereby reducing communication costs.}
\item 
   {We explore how to alleviate \textit{catastrophic forgetting} in FL while reducing communication costs. By optimizing local parameters only in the null space of global prompts, FedPR well preserves the previously acquired global knowledge in each round, maintaining competitive performance with only a few local data.}
\item 
   {We evaluate the performance of FedPR for federated MRI reconstruction. In comparison to the state-of-the-art FL methods, FedPR achieves superior performance in complex scenarios, \eg, less local data, lower communication costs, and faster convergence.}
   %non-i.i.d. 
%   {We analyze the benefits of learning a federated prompt for federated MRI reconstruction. We demonstrate that FedPR learns \textit{local knowledge} in the approximate \textit{null space} of the global prompt which protects the previously acquired global knowledge in each round, maintaining competitive performance with only a few local data (see \S\ref{sec:experiments}).}
\vspace*{-7pt}
\end{itemize}

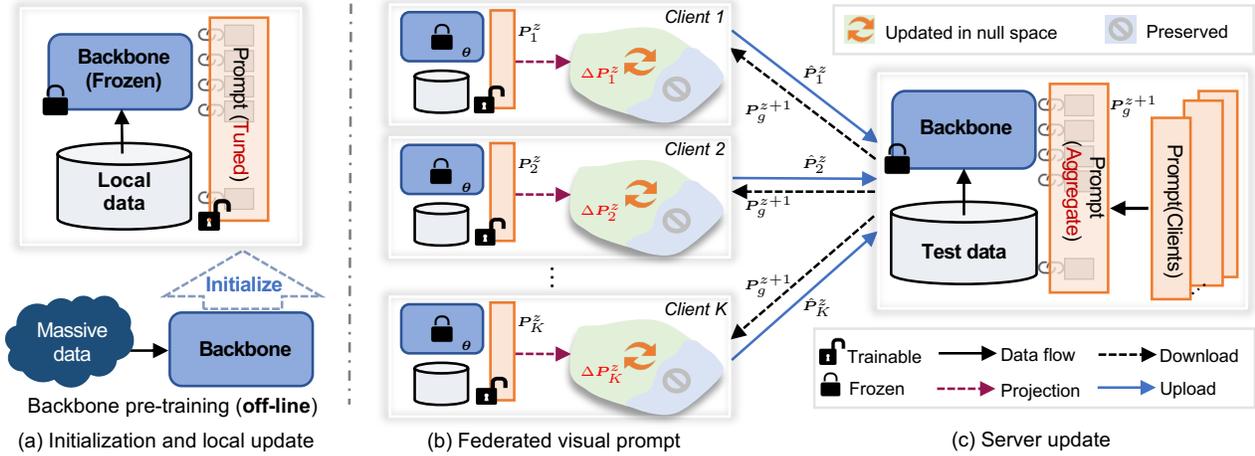
\begin{figure*}[!t]
    \vspace{-9pt}
	\begin{center}
		\includegraphics[width=0.96\linewidth]{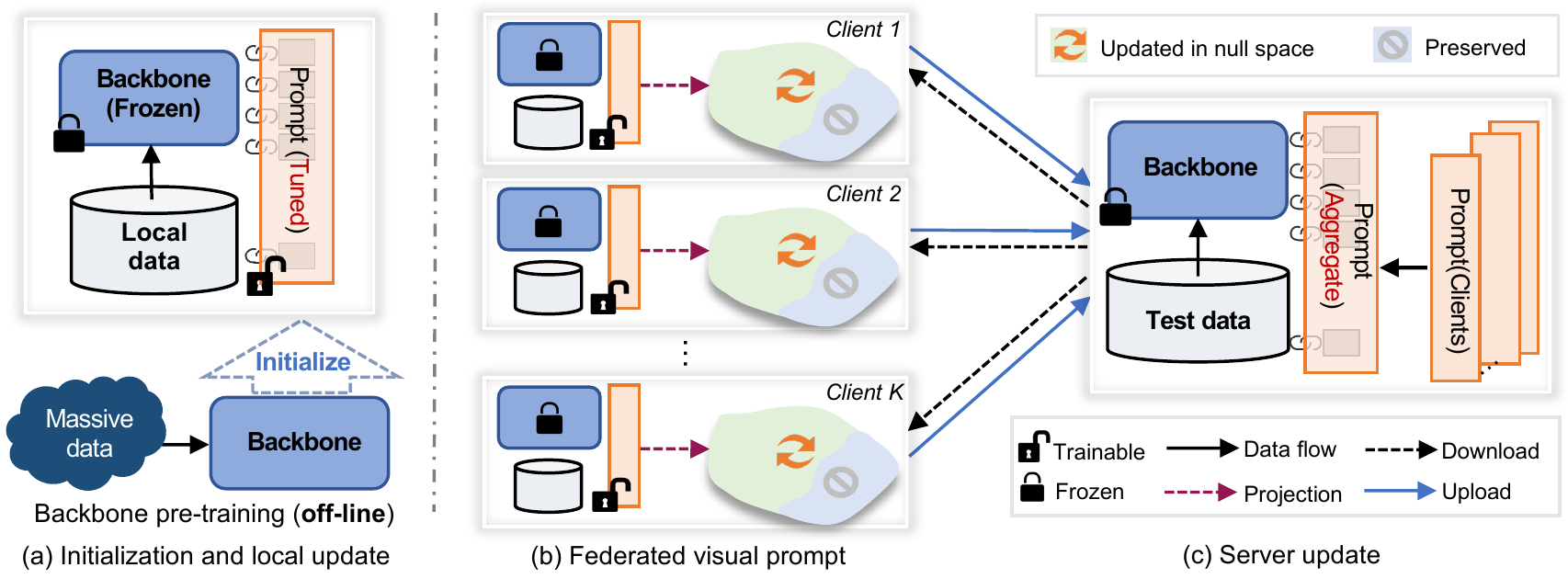}
		 \put(-284,160){ \tiny$\boldsymbol{\bm{P}}_{{1}}^{z}$} 
		 \put(-284,110){ \tiny$\boldsymbol{\bm{P}}_{{2}}^{z}$} 
		 \put(-284,50){ \tiny$\boldsymbol{\bm{P}}_{{K}}^{z}$} 
		 \put(-260,144){ \tiny${\color{red}\Delta \boldsymbol{\bm{P}}_{1}^z}$}
		 \put(-260,92){ \tiny${\color{red}\Delta \boldsymbol{\bm{P}}_{2}^z}$}
		 \put(-260,31){ \tiny${\color{red}\Delta \boldsymbol{\bm{P}}_{K}^z}$}
		 \put(-305,41){ \tiny$\boldsymbol{\theta}$}
		 \put(-305,101){ \tiny$\boldsymbol{\theta}$}
		 \put(-305,152){ \tiny$\boldsymbol{\theta}$}
		 \put(-60,132){ \tiny$\boldsymbol{\bm{P}}_{{g}}^{z+1}$}
		 \put(-176,144){ \tiny$\hat{\boldsymbol{\bm{P}}}_{{1}}^{z}$}
		 \put(-176,110){ \tiny$\hat{\boldsymbol{\bm{P}}}_{{2}}^{z}$}
		 \put(-176,56){ \tiny$\hat{\boldsymbol{\bm{P}}}_{{K}}^{z}$}
		 \put(-198,130){ \tiny${\boldsymbol{\bm{P}}}_{{g}}^{z+1}$}
		 \put(-198,94){ \tiny${\boldsymbol{\bm{P}}}_{{g}}^{z+1}$}
		 \put(-198,64){ \tiny${\boldsymbol{\bm{P}}}_{{g}}^{z+1}$}	
	\end{center}
    \vspace{-16pt}
	\captionsetup{font=small}
	\caption{\small\textbf{Illustration} of our \textbf{FedPR} method. (a) For each client, the backbone model is pre-trained on massive data and fine-tuned via prompt (see Sec.~\ref{sec:fedpr}). (b) \textit{Federated visual prompt} is executed by updating the local prompt of each client only in the approximate null space of global prompts while preserving the \textit{previously} acquired global knowledge (see Sec.~\ref{sec:null}). (c) Server update by aggregating local prompts.}
	\vspace{-18pt}
	\label{fig:2}
\end{figure*}

% \vspace{-3pt}
\section{Related Work}
\vspace{-2pt}
\noindent\textbf{Federated MRI Reconstruction.} MRI reconstruction refers to reconstructing images without aliasing artifacts from undersampled $k$-space data~\cite{wang2016accelerating,zhang2019reducing,feng2021donet,feng2021task,feng2021multi}. However, existing MRI reconstruction techniques are based on large-scale paired data, which is not only labor-intensive but also violates patient privacy~\cite{feng2022specificity}. Driven by these realistic problems, a few FL-based methods for MRI reconstruction have been proposed~\cite{guo2021multi,feng2022specificity,elmas2022federated}. Guo \textit{et al.} reduced data heterogeneity by iteratively aligning the data distribution between source and target clients~\cite{guo2021multi}. 
Feng \textit{et al.} proposed a personalized FL reconstruction scheme and introduced a weighted contrast regularization term to correct the update direction of global generalization~\cite{feng2022specificity}. 
% Elmas \textit{et al.} leveraged the prior information of MRI images to improve generalization and flexibility in multi-hospital collaboration~\cite{elmas2022federated}. 
However, these schemes either require frequent communication or are built on a large amount of local data.
As a result, these techniques will produce sub-optimal solutions for MRI reconstruction when there is \textit{limited bandwidth} and \textit{insufficient local training data}. In contrast to these works, we build a federated MRI model upon a strong \textit{pre-trained} model by updating and communicating only a \textit{small} portion of parameters for each client using \textit{limited} local data, achieving competitive results.
%current    

% \vspace{-2pt}
\noindent\textbf{Catastrophic Forgetting in FL.} 
% In continual learning, the network needs to continuously learn from a stream of data and train a series of tasks in sequence~\cite{de2021continual}. 
%
Due to the heterogeneous data distributions of different tasks in continual learning, when the network is fitted to the current task, the model parameters typically deviate from the areas where the previous knowledge is expected to be preserved~\cite{kim2018keep}. Therefore, catastrophic forgetting is a fundamental challenge of continual learning. Similarly, catastrophic forgetting also affects FL~\cite{lee2021preservation,xu2022acceleration,huang2022learn}. When the client receives the global model and continues to update locally, it will cause the global model to forget the knowledge of the previous round due to the data heterogeneity across clients~\cite{lee2021preservation}. As a result, existing FL techniques that address catastrophic forgetting are designed by referring to continual learning. For example, knowledge distillation-based FL methods maintain previously gained knowledge, but this method strongly relies on proxy data~\cite{huang2022learn,wei2022knowledge,usmanova2021distillation}; regularization-based FL methods regularize locally trained parameters, but they require finding a compromise between loss terms~\cite{shoham2019overcoming,xu2022acceleration,dong2022federated}. All these studies are built on classification tasks and cannot be directly extended to MRI reconstruction.
% , \eg, Lee \textit{et al.} kept the global perspective on locally available data only for the logits of the not-true classes~\cite{lee2021preservation,dong2022federated}.
Inspired by the success of null space in continual learning~\cite{wang2021training,kong2022balancing,lin2022towards}, we try to update the \textit{local} prompt in the approximate \textit{null space} of global prompts, thereby preserving the knowledge acquired by the global model in the previous round and avoiding \textit{catastrophic forgetting}.

% In continual learning, the network needs to continuously learn from a stream data and train a series of tasks in sequence~\cite{de2021continual}. Due to the heterogeneous data distribution of each task, when the network is fitted to the current task, the model parameters typically deviate from the areas where the previous knowledge is expected to be preserved~\cite{kim2018keep}. Therefore, catastrophic forgetting is the biggest challenge of continual learning. Similarly, catastrophic forgetting also affects FL~\cite{lee2021preservation,xu2022acceleration,huang2022learn}. When the client receives the global model and continues to update locally, it will cause the global model to forget the knowledge of the previous round due to the data heterogeneity across clients~\cite{lee2021preservation}. Therefore, we can consider how to solve the catastrophic forgetting of FL and accelerate convergence from the perspective of continual learning. In this paper, we try to update the \textit{local} prompt in the approximate \textit{null space} of the feature covariance matrix of the global prompt, to preserve the knowledge acquired by the global model in the previous round and avoid \textit{catastrophic forgetting}.

% \vspace{-2pt}
\noindent\textbf{Prompt Learning.} Prompt tuning was initially proposed to enable pre-trained language models to ``understand" downstream tasks by adding trainable tokens to the input text~\cite{liu2021pre}. For example, Jia \textit{et al.} achieved good performance on downstream tasks by introducing a small number of learnable tokens into the input space and keeping the backbone frozen~\cite{jia2022visual}. Hyojin \textit{et al.} suggested to modify pixel space for the frozen visual models~\cite{sssss}. Nie \textit{et al.} proposed a unified prompt tuning framework that performs on different CNN and transformer-based architectures by training only a few additional parameters~\cite{nie2022pro}. Although these works have made progress on various visual tasks, prompt is still limited to centralized systems, and the effectiveness of prompt in distributed framework remains uninvestigated. 
Therefore, this work focuses on how to learn a \textit{federated visual prompt} to effectively tackle the three key problems in federated MRI reconstruction, \ie, the issues \myhyperlink{Q1}{\ding{182}}-\myhyperlink{Q3}{\ding{184}}.

\vspace{-3pt}
\section{Methodology}\label{sec:method}
\vspace{-2pt}
% Before detailing our proposed algorithm FedPR for MRI reconstruction (\S\ref{sec:fedpr} and \S\ref{sec:null}), we first introduce the notion of federated MRI (\S\ref{sec:federatedmri}).
\vspace{-2pt}
\subsection{Federated MRI Reconstruction}\label{sec:federatedmri}
\vspace{-3pt}
MRI reconstruction is an inverse problem of recovering an artifact-free image $\mathbf{y}$ from its undersampled observation $\mathbf{x}$ that can reduce the online collection time and improve the patient experience~\cite{wang2016accelerating,yang2018admm,sun2019deep}. Formally, such undersampling process can be expressed as
\vspace{-5pt}
\begin{equation} 
\mathbf{x}=\mathcal{F}^{-1}\left(\mathcal{M} \odot \mathcal{F}(\mathbf{y})+\epsilon\right),
\label{eq:ds}
\vspace{-4pt}
\end{equation}
where $\mathcal{F}$ is multi-dimensional Fourier transform, $\epsilon$ is the measurement noise, and $\mathcal{M}$ is the binary mask operator to undersample the data points in the Fourier space. 
% Given $N$ training samples, we denote $\mathbf{x}$ represent the undersampled input and $\mathbf{y}$ be the fully sampled data to be recovered. A centralized reconstruction process can be expressed as:
% \vskip -10pt
% \begin{equation}
% \mathcal{L}_{rec}=\frac{1}{N} \sum_{n=1}^{N}\left\|f\left(\mathbf{x};\boldsymbol{w}\right)-\mathbf{y}\right\|_{1},
% \label{eq:2}
% \end{equation}
% where $f\left(\mathbf{x}\right)$ is the convolutional network parameterized by $\boldsymbol{w}$. 

The centralized approach violates patient privacy protection roles because it requires collecting large amounts of training data from different hospitals. 
As a remedy, Federated MRI has been suggested by deploying MRI reconstruction in a distributed manner~\cite{feng2022specificity,guo2021multi}. 
Suppose there are $K$ hospitals (local clients). 
The private data of all hospitals can be expressed as $\mathcal{D} = \left\{\mathcal{D}^{1}, \mathcal{D}^{2}, \ldots, \mathcal{D}^{K}\right\}$, where each contains limited pairs of undersampled samples $\mathbf{x}_i$ and fully-sampled images $\mathbf{y}_i$. Federated MRI aims to learn a global model from the whole dataset $\mathcal{D}$ in a distributed manner, which can be described as 
% For each client, we train a local model $f$ by minimizing the following loss:
\vskip -10pt
\begin{equation}
{\arg \underset{\boldsymbol{w}}\min } \mathcal{L}(\boldsymbol{w})= \sum_{k=1}^K \frac{|\mathcal{D}^k|} {|\mathcal{D}|} \mathcal{L}_k(\boldsymbol{w}),
\label{eq:1}
\vspace{-5pt}
\end{equation}
where $|\mathcal{D}|$ denotes the number of samples in $\mathcal{D}$, and  $\mathcal{L}_k(\boldsymbol{w})$ is the empirical loss of client $k$,
\vspace{-8pt}
\begin{equation}
\mathcal{L}_{k}(\boldsymbol{w}_{{}})=\mathbf{E}_{\left(\mathbf{x}, \mathbf{y}\right) \in \mathcal{D}^k} \ell_k\left(f\left(\mathbf{x} ;\boldsymbol{w}\right)\right),
\vspace{-6pt}
\end{equation}
where $\ell_k$ represents the local loss for MRI reconstruction, \eg, $L_1$ loss. After training, the reconstructed image $\hat{\mathbf{y}}$ can be produced by $f\left(\mathbf{x} ;\boldsymbol{w}\right)$. 
% In each round $z=1,...,Z$, each clients update with learning rate $\eta_{k}$ by the following update rules:
% \vskip -8pt
% \begin{equation}
% \boldsymbol{w}_{{k}}^{z,t+1} \leftarrow \boldsymbol{w}_{{k}}^{z,t}-\eta_{k} \nabla \ell_k\left(\mathbf{x}^{k};\boldsymbol{w}_{{k}}^{z,t+1}\right),
% \vspace{-6pt}
% \end{equation}
% where $t$ represents the number of local updates. After completing a round of local updates, all participating clients send their updated weights $\boldsymbol{w}_{{k}}^{z}$ to the server performing aggregation. Such process can be expressed as follows
% \vskip -10pt
% \begin{equation}
% \boldsymbol{w}_{{g}}^{z+1}\leftarrow \sum_{k=1}^K\frac{\mathcal{D}^k}{\mathcal{D}} \boldsymbol{w}_{{k}}^{z},
% \vspace{-6pt}
% \end{equation}
% where $\boldsymbol{w}_{{g}}^{z+1}$ represents the global parameters of round $z+1$.

\vspace{-2pt}
\subsection{Learning Federated Prompt}\label{sec:fedpr}
\vspace{-2pt}
% As we mentioned before, current federated MRI requires communication of the entire or most of the model parameters with the server, resulting in significant communication cost. 
% With this perspective, 
As mentioned in Sec.~\ref{sec:intro}, there are three key issues, \ie, \myhyperlink{Q1}{\ding{182}}-\myhyperlink{Q3}{\ding{184}}, with the existing federated MRI algorithms. To tackle the issue \myhyperlink{Q2}{\ding{183}}, we freeze the backbone of the pre-trained model while only tuning and communicating a few learnable parameters for the clients and server. 
More importantly, it enables the model to achieve very competitive results on limited local data, thereby providing an effective solution to issue \myhyperlink{Q1}{\ding{182}}.

As shown in Fig.~\ref{fig:2}, given a pre-trained model with parameters $\boldsymbol{\theta}$, we introduce a set of continuous embeddings $\bm{P}=\left\{\bm{p}_1, \bm{p}_2, \cdots, \bm{p}_l\right\}$ as the prompts in the input space of each layer~\cite{jia2022visual}, where $l$ is the number of prompts. Thus, the overall parameters can be expressed as $\boldsymbol{w}\!=\!\{ \bm{P}, \boldsymbol{\theta} \}$. 
Suppose there are $Z$ rounds of communication, each round contains $T$ local updates. During the local update, for the $k$-th client, only the client-specific prompts $\bm{P}_k$ are learnable and communicated while the backbone network is frozen. Therefore, Eq.~\eqref{eq:1} can be rewritten as:   
\vskip -12pt
\begin{equation}
\bm{P}_k = {\arg}~\underset{\bm{P}}{\min } \mathcal{L}(\bm{P})= \sum_{k=1}^K \frac{|\mathcal{D}^k|}{|\mathcal{D}|} \mathcal{L}_k(\bm{P}),
\label{eq:5}
\vspace{-4pt}
\end{equation}
where the empirical loss for prompt tuning of client $k$ can be expressed as 
\vspace{-7pt}
\begin{equation}
\mathcal{L}_{k}(\boldsymbol{\bm{P}}_{{}})=\mathbf{E}_{\left(\mathbf{x}, \mathbf{y}\right) \in \mathcal{D}^k} \ell_k\left(f\left(\mathbf{x}; \bm{P},\boldsymbol{\theta}\right)\right).
\vspace{-4pt}
\end{equation}
Benefited from pre-trained models, federated prompt tuning can be both parameter- and communication-efficient and effective in achieving competitive results with a small amount of local data, thereby serving as a favorable solution to the issues \myhyperlink{Q1}{\ding{182}} and \myhyperlink{Q2}{\ding{183}}.

\noindent{\textbf{Local Update Step:}} In each communication round $z=\left\{1,2,...,Z\right\}$, the clients are optimized using the following update rules with a learning rate of $\eta_k$:
\vskip -8pt
\begin{equation}\label{eq:6}
\boldsymbol{\bm{P}}_{{k}}^{z,t+1} \leftarrow \boldsymbol{\bm{P}}_{{k}}^{z,t}-\eta_{k} \nabla \ell_k\left(\mathbf{x}^{k};\boldsymbol{\bm{P}}_{{k}}^{z,t}\right),
\vspace{-3pt}
\end{equation}
where $t$ denotes the $t$-th update of the local clients. 

\noindent{\textbf{Server Update Step:}} After a round of  local updates, all participating clients send their updated prompts $\boldsymbol{\bm{P}}_{{k}}^{z}$ to the server performing aggregation. Such process can be expressed as follows
\vskip -15pt
\begin{equation}\label{eq7}
\boldsymbol{\bm{P}}_{{g}}^{z+1}\leftarrow \sum_{k=1}^K\frac{|\mathcal{D}^k|}{|\mathcal{D}|} \boldsymbol{\bm{P}}_{{k}}^{z},
\vspace{-4pt}
\end{equation}
where $\boldsymbol{\bm{P}}_{{g}}^{z+1}$ denotes the global prompts of round $z+1$. 
To sum up, the issue \myhyperlink{Q2}{\ding{183}} can be largely mitigated because there are only a small number of learnable parameters $\bm{P}$ communicated between the server and local clients. After $Z$ rounds of communication, we can get a robust global model parameterized by $\boldsymbol{\bm{P}}_{{g}}$ without sharing local private data.

\begin{figure}[t]
    \vspace{-8pt}
	\begin{center}
		\includegraphics[width=\linewidth]{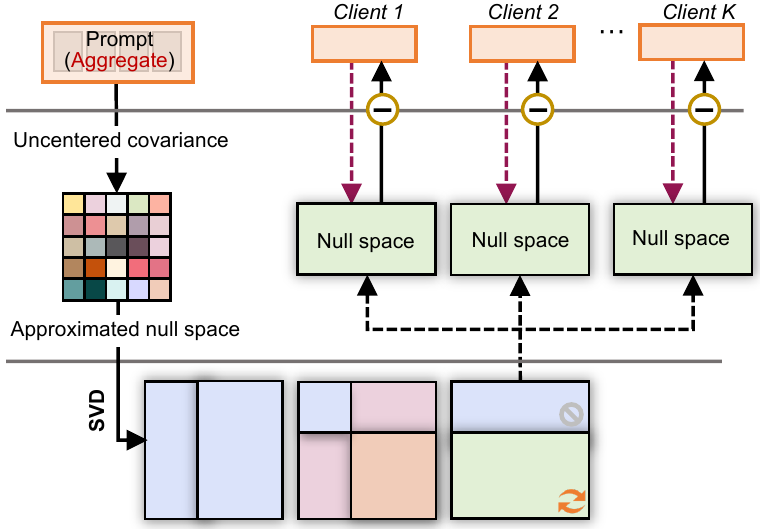}
		\put(-195,24){ \tiny$U_1^{z+1}$} 
		\put(-173,24){ \tiny$U_2^{z+1}$} 
		\put(-147,35){ \tiny$\Lambda_1^{z+1}$} 
		\put(-125,14){ \tiny$\Lambda_2^{z+1}$} 
		\put(-95,36){ \tiny$\left(U_1^{z+1}\right)\!^{\top}$}
		\put(-95,15){ \tiny$\left(U_2^{z+1}\right)\!^{\top}$} 
		 \put(-142,118){ \tiny$\boldsymbol{\bm{P}}_{{1}}^{z}$} 
		 \put(-94,118){ \tiny$\boldsymbol{\bm{P}}_{{2}}^{z}$} 
		 \put(-43,118){ \tiny$\boldsymbol{\bm{P}}_{{K}}^{z}$} 
		 \put(-117,136){ \tiny$\hat{\boldsymbol{\bm{P}}}_{{1}}^{z}$}
		 \put(-69,136){ \tiny$\hat{\boldsymbol{\bm{P}}}_{{2}}^{z}$}
		 \put(-17,136){ \tiny$\hat{\boldsymbol{\bm{P}}}_{{K}}^{z}$}	
		 \put(-118,115){ \tiny${\color{red}\Delta \boldsymbol{\bm{P}}_{1}^z}$}
		 \put(-70,115){ \tiny${\color{red}\Delta \boldsymbol{\bm{P}}_{2}^z}$}
		 \put(-18,115){ \tiny${\color{red}\Delta \boldsymbol{\bm{P}}_{K}^z}$}	
		 \put(-240,85){ \tiny$\boldsymbol{\Sigma}_{g}^{z+1}$}  
		 
	\end{center}
	\vspace{-16pt}
	\captionsetup{font=small}
	\caption{\small\textbf{Illustration} of \textbf{federated visual prompt} for updating local prompts in the null space of global prompts.}
	\vspace{-13pt}
	\label{fig:3}
\end{figure}
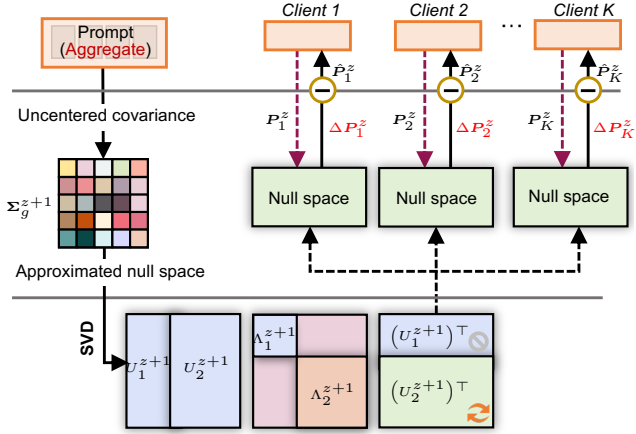

\begin{algorithm}[!t]
    \caption{FedPR}
    \label{alg:fedpr}
    \KwIn{Private datasets from $K$ clients: $\mathcal{D}^{1}, \mathcal{D}^{2}, \ldots, \mathcal{D}^{K}$, local updates $T$, communication rounds $Z$, pre-trained model parameters $\boldsymbol{\theta}$, prompt embeddings $\bm{P}$, learning rate $\eta_{}$, hyperparameter $\gamma$;}
% \KwOut{The final global model $\boldsymbol{w}_{G}$}
// \textbf{ServerExecution:}\\
Initialize global prompt $\bm{P}_{g}$ with parameters $\boldsymbol{\theta}$.

\For {each communication round $z\in\left\{1,2,...,Z\right\}$}{
    \For{each client $k\in\left\{1,2,...K\right\}$ in parallel}{
             $\boldsymbol{\bm{P}}_{{k}}^{z}\leftarrow \boldsymbol{\bm{P}}_{{g}}^{z}$;\\
             $\hat{\boldsymbol{\bm{P}}}_{{k}}^{z}\leftarrow \text{LocalUpdate}(k,\boldsymbol{\bm{P}}_{{k}}^{z})$;
    }
    $\boldsymbol{\bm{P}}_{{g}}^{z+1}\leftarrow \sum_{k=1}^K\frac{|\mathcal{D}^k|}{|\mathcal{D}|} \hat{\boldsymbol{\bm{P}}}_{{k}}^{z}$;\\
    Compute the uncentered covariance matrix:\\
    $\boldsymbol{\Sigma}_{g}^{z+1}=\left(\boldsymbol{\bm{P}}_{g}^{z+1}\right)\!^{\top} \boldsymbol{\bm{P}}_{g}^{z+1}$;\\
     Approximate the null space of $\boldsymbol{\Sigma}_{g}^{z+1}$:\\
     $\boldsymbol{\Sigma}_{g}^{z+1} = \bm{U}^{z+1} \boldsymbol{\Lambda}^{z+1} \left(\bm{U}^{z+1}\right)\!^{\top}$;\\
     Select the smallest diagonal singular values of $\bm{\Lambda}_2^{z+1}$ with the ratio of $\gamma$;\\
     Get $\bm{U}_2^{z+1}$ corresponding to $\bm{\Lambda}_2^{z+1}$.
}
return ${\boldsymbol{\bm{P}}}_{{g}}^{z+1}$ 

// \textbf{LocalUpdate} ($k$, $\bm{P}_{k}$)\textbf{:}

    \For{each local epoch $t\in\left\{1,2,...T\right\}$}
            {
                      Get the updated parameters $\Delta \boldsymbol{\bm{P}}_{k}^z$:\\
     $\Delta \boldsymbol{\bm{P}}_{k}^z= \bm{U}_2^{z+1}\left(\bm{U}_2^{z+1}\right)\!^{\top} \boldsymbol{\bm{P}}_{{k}}^{z}$;\\
                 
                  $\hat{\boldsymbol{\bm{P}}}_{{k}}^{z,t+1}\leftarrow {\boldsymbol{\bm{P}}}_{{k}}^{z,t}\ - \eta_k \Delta\boldsymbol{\bm{P}}_{{k}}^{z,t}$\;

}
return $\hat{\boldsymbol{\bm{P}}}_{{k}}^{z}$
% \vspace{-4pt}
\end{algorithm}
% \vskip -25pt
% \vspace{-32pt}

\subsection{Local Updating in Prompting Null Space}\label{sec:null}
\vspace{-4pt}
The local update mechanism in Eq.~(\ref{eq:6}), however, suffers from the issue \myhyperlink{Q3}{\ding{184}}, \ie, catastrophic forgetting, due to data heterogeneity of different clients.
Inspired by the continual learning method Adam-NSC~\cite{wang2021training} that sequentially optimizes network parameters in the null space at  feature level, we suggest to optimize the local model in the null space of global prompts. Specifically, we train the local model in the approximate null space of the global prompts in a distributed way, preventing knowledge outside the local distribution from being overwritten and resulting in catastrophic forgetting. In comparison to~\cite{wang2021training}, approximating the null space of global prompts instead of features does not require a large amount of local data and not increase the communication cost, thus offering a feasible way to tackle the three issues \myhyperlink{Q1}{\ding{182}}-\myhyperlink{Q3}{\ding{184}} in federated MRI.

As shown in Fig.~\ref{fig:3}, we first compute the uncentered covariance matrix of $z\!+\!1$ round global prompt  $\boldsymbol{\bm{P}}_{{g}}^{z+1}$ by
\vspace{-6pt}
\begin{equation}
\boldsymbol{\Sigma}_{g}^{z+1}=\left(\boldsymbol{\bm{P}}_{g}^{z+1}\right)\!^{\top} \boldsymbol{\bm{P}}_{g}^{z+1}.
\vspace{-6pt}
\end{equation}
Then, we find the approximate null space of $\boldsymbol{\Sigma}_{g}^{z+1}$ by apply SVD to it
\vskip -10pt
\begin{equation}
\boldsymbol{\Sigma}_{g}^{z+1} = \bm{U}^{z+1} \boldsymbol{\Lambda}^{z+1} \left(\bm{U}^{z+1}\right)\!^{\top},
%=\operatorname{SVD}\left(\mathcal{P}_{g}^{z+1}\right),
\vspace{-9pt}
\end{equation}
where
\vskip -20pt
\begin{equation}
\begin{aligned}
\\&~~~ \bm{U}^{z+1}\!=\!\left[\bm{U}_1^{z+1}, \bm{U}_2^{z+1}\right],
\\&~~~ \boldsymbol{\Lambda}^{z+1}\!=\!\left[\begin{array}{cc}\boldsymbol{\Lambda}_1^{z+1} & 0 \\ 0 & \boldsymbol{\Lambda}_2^{z+1}\end{array}\right].
\end{aligned}
\vspace{-5pt}
\end{equation}
Here $\bm{U}_1^{z+1}$ are the singular vectors corresponding to the large singular values in $\boldsymbol{\Lambda}_1^{z+1}$~\cite{wang2021training}.
According to principal component analysis (PCA), $\bm{U}_1^{z+1}$ can be considered as principal components, which contain the \textit{global knowledge} that has been captured in the previous round. Thus, we have $\boldsymbol{\Sigma}_{g}^{z+1} \approx \bm{U_}1^{z+1} \boldsymbol{\Lambda}_1^{z+1}\left(\bm{U}_1^{z+1}\right)\!^{\top}$.
%
%thereby $\boldsymbol{\Lambda}_{g}^{z+1} \bm{U}_2^{z+1} \approx \bm{U}_1^{z+1} \boldsymbol{\Lambda}_1^{z+1}\left(\bm{U}_1^{z+1}\right)\!^{\top} \bm{U}_2^{z+1}\!=\!0$. 
%
And the range space of $\bm{U}_2^{z+1}$ can be denoted as the approximate null space of $\boldsymbol{\Sigma}_{g}^{z+1}$~\cite{wang2021training}. 
%, where 
In our implementation, we introduce a $\gamma\%$ to control $\bm{U}_2^{z+1}$ by selecting the last $\gamma\%$ of singular values in $\boldsymbol{\Lambda}_2^{z+1}$ to constitute $\bm{U}_2^{z+1}$.

To tackle the issue \myhyperlink{Q3}{\ding{184}}, $\boldsymbol{\bm{P}}_{{k}}^{z}$ can be projected into the approximate null space of $\boldsymbol{\Sigma}_{g}^{z+1}$ to get the updated parameters, while the principal components $\bm{U}_1^{z+1}$ containing the \textit{global knowledge} from the previous round are \textit{preserved}. Thus, the updated parameters $\Delta \boldsymbol{\bm{P}}_{k}^z$ can be optimized in the null space of $\boldsymbol{\Sigma}_{g}^{z+1}$ by
\vskip -8pt
\begin{equation}
\Delta \boldsymbol{\bm{P}}_{k}^z= \bm{U}_2^{z+1}\left(\bm{U}_2^{z+1}\right)\!^{\top} \boldsymbol{\bm{P}}_{{k}}^{z},
\vspace{-2pt}
\end{equation}
where $\bm{U}_2^{t+1}\left(\bm{U}_2^{t+1}\right)\!^{\top}$ is the projection operator~\cite{meyer2000matrix}. Formally, the local update step of Eq.~\eqref{eq:6} can be rewritten as
\vskip -6pt
\begin{equation}
\hat{\boldsymbol{\bm{P}}}_{{k}}^{z,t+1}\leftarrow {\boldsymbol{\bm{P}}}_{{k}}^{z,t}\ - \eta_k \Delta
\boldsymbol{\bm{P}}_{{k}}^{z,t}.
\vspace{-3pt}
\end{equation}
The detailed FedPR algorithm is given in Algorithm~\ref{alg:fedpr}.

\section{Experiments}
% \vspace{-4pt}
\subsection{Experimental Setup}
\vspace{-4pt}
\noindent{\textbf{Implementation Details.}} Our method is trained by Pytorch with one NVIDIA Tesla V$100$ GPU and $32$GB of memory. We use Adam as the optimizer, with a momentum of $0.9$ and weight decay of $0.0005$. The models are trained with $50$ communication rounds and $10$ local epochs for each round. We set the batch size and initial learning rate to $8$ and $1 \times 10^{-1}$, respectively. The hyperparameter $\gamma$ is empirically set to $80$\%. The prompt embeddings are added with a size of $8\times20\times256$.

% \vspace{-13pt}
\noindent{\textbf{Datasets.}} We use \textbf{fastMRI}\footnote{\url{https://fastmri.org/}.}, the largest publicly available MRI dataset, to train the pre-training model, with Swin Transformers and a standard ConvNet head serving as the backbone of our experiments~\cite{liu2021swin}.
We note that the mean and std of BN layer in the head are also shared with the server to update the original statistical properties. For the clients, we employ four datasets with a size of $320\times320$, including \textbf{FeTS}\footnote{\url{https://www.synapse.org/\#!Synapse:syn28546456/wiki/}.}~\cite{pati2022federated}, \textbf{IXI}\footnote{\url{https://brain-development.org/}}~\cite{ixi}, and two clinical datasets, to comprise our local data. In our experiments, we divide them into $15$ clients according to the acquiring institutions, \ie, \textbf{FeTS} is divided into $10$ clients with only $120$, $304$, $304$, $240$, $160$, $160$, $280$, $224$, $264$, and $184$ images, respectively; \textbf{IXI} is divided into $3$ clients with only $360$, $328$, and $296$ images, respectively; two clinical datasets collected from the United Imaging Healthcare uMR  $790$ scanner and 3T Siemens Magnetom Skyra system form two clients with $432$ and $360$ images, respectively. The input data of each client employs a \textit{1D random} sampling pattern with \textit{3$\times$} acceleration. We note that our clients have only a small number of images. In comparison, FedMRI~\cite{feng2022specificity} and FL-MRCM~\cite{guo2021multi} require thousands to tens of thousands of images. For \texttt{In-Federation} evaluation, we divide each dataset with a ratio of $7:3$ for local \texttt{train/test}. For \texttt{Out-of-Federation} evaluation, we use the test set of \textbf{FeTS} as our \texttt{test} set because it comes from a separate institution with a quite different distribution.

% \vspace{-4pt}
\noindent{\textbf{Baselines.}}~To demonstrate the effectiveness of our proposed method, we compare it with three categories of methods, including: \textbf{a)} SingleSet, a single model that each client is trained with their local data without FL; \textbf{b)} Centralized, a single model that is trained with the combination of all the local data, which serves as the upper-bound of FL models; and \textbf{c)} eight state-of-the-art FL algorithms, including: (1) FedAvg~\cite{mcmahan2017communication}, a classical FL algorithm that is trained by averaging parameters of all the participating clients; (2) FedBN~\cite{li2021fedbn}, a FL algorithm that alleviates the client-shift by using  batch normalization on each local client; (3) FedProx~\cite{li2020federated}, a FL algorithm that applies a proximal term to the local objective function; (4) SCAFFOLD~\cite{karimireddy2020scaffold}, a FL algorithm that uses control variates to correct the client-drift; (5) MOON~\cite{li2021model}, a FL algorithm that utilizes the similarity between model representations to correct the local update; (6) FedReg~\cite{xu2022acceleration}, a FL algorithm that regularizes locally trained parameters with the loss on generated pseudo data to alleviate knowledge forgetting; (7) FL-MRCM~\cite{guo2021multi}, a federated MRI algorithm that aligns the latent features between the source and target clients; and (8) FedMRI~\cite{feng2022specificity}, a federated MRI algorithm that preserves the client-specific properties to improve the FL performance. For a fair comparison, we adopt Swin Transformers and a standard ConvNet head as the reconstruction network for each client among all the baselines.

\begin{table*}[t]
\renewcommand{\arraystretch}{1.3}
	\caption{\small \textbf{Quantitative comparison} of state-of-the-art FL methods with regard to \texttt{In-Federation} and \texttt{Out-of-Federation} scenarios, where \textbf{\# Com.cost} is the communication cost, Ub indicates the upper-bound of FL algorithms, ${\color{ForestGreen}\uparrow}$ and ${\color{red}\downarrow}$ indicate increments and decrements compared with FedAvg (\textit{w.}~FFt). Detailed analyses are provided in Sec.~\ref{sec:acc}.}
	\vspace{-10pt}
	\label{tab:1}

	\fontsize{9}{9}\selectfont
	\centering
	\begin{tabular}{l c cc cc cc cc cc cc}
\toprule
\multicolumn{1}{l}{}&&\multicolumn{6}{c}{\textbf{\texttt{In-Federation}}}&\multicolumn{6}{c}{\textbf{\texttt{Out-of-Federation}}} \\%\midrule
\cmidrule(lr){3-8} \cmidrule(l){9-14}
\textbf{Method}&\multicolumn{1}{c}{\textbf{\# Com.cost}}&\multicolumn{2}{c}{\textbf{PSNR}} &\multicolumn{2}{c}{\textbf{SSIM}} &\multicolumn{2}{c}{\textbf{NMSE}}&\multicolumn{2}{c}{\textbf{PSNR}} &\multicolumn{2}{c}{\textbf{SSIM}} &\multicolumn{2}{c}{\textbf{NMSE}}\\

\cmidrule(r){1-1} \cmidrule{2-2} \cmidrule(lr){3-4} \cmidrule(lr){5-6} \cmidrule(lr){7-8} \cmidrule(lr){9-10} \cmidrule(lr){11-12} \cmidrule(l){13-14} 
%\noalign{\smallskip}
%5.0	0.5	0.4	0.9	0.9	0.5	0.9	1.0
SingleSet &\multicolumn{1}{|c|}{$18.43$ M}  &\multicolumn{2}{c}{28.79\stdvd{$3.35$}} &\multicolumn{2}{c}{0.805\stdvd{$0.099$}}&\multicolumn{2}{c}{0.021\stdvd{$0.011$}} &\multicolumn{2}{|c}{28.92\stdvd{$2.15$}} &\multicolumn{2}{c}{0.808\stdvd{$0.091$}}&\multicolumn{2}{c}{0.032\stdvd{$0.011$}}\\
Centralized (Ub)  &\multicolumn{1}{|c|}{$18.43$ M} &\multicolumn{2}{c}{36.71\stdvu{$4.57$}} &\multicolumn{2}{c}{0.947\stdvu{$0.044$}}&\multicolumn{2}{c}{0.009\stdvu{$0.005$}} &\multicolumn{2}{|c}{35.73\stdvu{$4.66$}} &\multicolumn{2}{c}{0.940\stdvu{$0.041$}}&\multicolumn{2}{c}{0.008\stdvu{$0.013$}}\\\hline
FedAvg (\textit{w.}~FFt)~\cite{mcmahan2017communication} &\multicolumn{1}{|c|}{$18.43$ M}  &\multicolumn{2}{c}{32.14\stdvw{$0.00$}} &\multicolumn{2}{c}{0.903\stdvw{$0.000$}}&\multicolumn{2}{c}{0.010\stdvw{$0.000$}} &\multicolumn{2}{|c}{31.07\stdvw{$0.00$}} &\multicolumn{2}{c}{0.899\stdvw{$0.000$}}&\multicolumn{2}{c}{0.021\stdvw{$0.000$}}\\
FedBN~\cite{li2021fedbn} &\multicolumn{1}{|c|}{$18.40$ M} &\multicolumn{2}{c}{28.31\stdvd{$3.83$}} &\multicolumn{2}{c}{0.817\stdvd{$0.086$}}&\multicolumn{2}{c}{0.044\stdvd{$0.034$}} &\multicolumn{2}{|c}{25.65\stdvd{$5.42$}} &\multicolumn{2}{c}{0.731\stdvd{$0.186$}}&\multicolumn{2}{c}{0.080\stdvd{$0.059$}}\\
FedProx~\cite{li2020federated} &\multicolumn{1}{|c|}{$18.43$ M}  &\multicolumn{2}{c}{32.84\stdvu{$0.70$}} &\multicolumn{2}{c}{0.901\stdvu{$0.002$}}&\multicolumn{2}{c}{0.010\stdvw{$0.000$}}&\multicolumn{2}{|c}{31.98\stdvu{$0.91$}} &\multicolumn{2}{c}{0.905\stdvu{$0.006$}}&\multicolumn{2}{c}{0.018\stdvu{$0.003$}}\\
SCAFFOLD~\cite{karimireddy2020scaffold} &\multicolumn{1}{|c|}{$18.43$ M} &\multicolumn{2}{c}{33.08\stdvu{$0.94$}} &\multicolumn{2}{c}{0.914\stdvu{$0.010$}}&\multicolumn{2}{c}{0.009\stdvu{$0.001$}}&\multicolumn{2}{|c}{32.20\stdvu{$1.13$}} &\multicolumn{2}{c}{0.915\stdvu{$0.016$}}&\multicolumn{2}{c}{0.018\stdvu{$0.003$}}\\
MOON~\cite{li2021model} &\multicolumn{1}{|c|}{$18.43$ M} &\multicolumn{2}{c}{34.06\stdvu{$1.92$}} &\multicolumn{2}{c}{0.927\stdvu{$0.023$}}&\multicolumn{2}{c}{0.008\stdvu{$0.002$}}&\multicolumn{2}{|c}{31.16\stdvu{$0.09$}} &\multicolumn{2}{c}{0.907\stdvu{$0.008$}}&\multicolumn{2}{c}{0.023\stdvd{$0.002$}}\\
FedReg~\cite{xu2022acceleration} &\multicolumn{1}{|c|}{$18.43$ M} &\multicolumn{2}{c}{33.29\stdvu{$1.15$}} &\multicolumn{2}{c}{0.890\stdvu{$0.013$}}&\multicolumn{2}{c}{0.009\stdvu{$0.001$}}&\multicolumn{2}{|c}{32.41\stdvu{$1.34$}} &\multicolumn{2}{c}{0.907\stdvu{$0.008$}}&\multicolumn{2}{c}{0.017\stdvu{$0.004$}}\\
FL-MRCM~\cite{guo2021multi} &\multicolumn{1}{|c|}{$18.43$ M} &\multicolumn{2}{c}{33.60\stdvu{$1.46$}} &\multicolumn{2}{c}{0.922\stdvu{$0.019$}}&\multicolumn{2}{c}{0.013\stdvd{$0.003$}}&\multicolumn{2}{|c}{32.72\stdvu{$1.65$}} &\multicolumn{2}{c}{0.911\stdvu{$0.012$}}&\multicolumn{2}{c}{0.016\stdvu{$0.005$}}\\
FedMRI~\cite{feng2022specificity} &\multicolumn{1}{|c|}{$17.46$ M}&\multicolumn{2}{c}{33.35\stdvu{$1.21$}} &\multicolumn{2}{c}{0.923\stdvu{$0.020$}}&\multicolumn{2}{c}{0.014\stdvd{$0.004$}}&\multicolumn{2}{|c}{32.00\stdvu{$0.93$}} &\multicolumn{2}{c}{0.914\stdvu{$0.015$}}&\multicolumn{2}{c}{0.019\stdvu{$0.002$}}\\
{\cellcolor{mypink}$~\textbf{\texttt{Ours}}$ (\textit{w.}~Pro)} & \multicolumn{1}{|c|}{{\cellcolor{mypink}$\textbf{0.11}$ M}} &\multicolumn{2}{c}{{\cellcolor{mypink}35.29\stdvu{$3.15$}}} &\multicolumn{2}{c}{{\cellcolor{mypink}0.927\stdvu{$0.024$}}}&\multicolumn{2}{c}{{\cellcolor{mypink}0.009\stdvu{$0.001$}}}&\multicolumn{2}{|c}{{\cellcolor{mypink}34.65\stdvu{$3.58$}}} &\multicolumn{2}{c}{{\cellcolor{mypink}0.921\stdvu{$0.022$}}}&\multicolumn{2}{c}{{\cellcolor{mypink}0.010\stdvu{$0.011$}}}\\
{\cellcolor{mypink}$~\textbf{\texttt{Ours}}$ (Full)} &\multicolumn{1}{|c|}{{\cellcolor{mypink}$\textbf{0.11}$ M}} &\multicolumn{2}{c}{{\cellcolor{mypink}\textbf{36.43}\stdvu{$\underline{4.30}$}}} &\multicolumn{2}{c}{{\cellcolor{mypink}\textbf{0.945}\stdvu{$\underline{0.042}$}}}&\multicolumn{2}{c}{{\cellcolor{mypink}\textbf{0.007}\stdvu{$\underline{0.003}$}}}&\multicolumn{2}{|c}{{\cellcolor{mypink}\textbf{35.60}\stdvu{$\underline{4.53}$}}} &\multicolumn{2}{c}{{\cellcolor{mypink}\textbf{0.939}\stdvu{$\underline{0.040}$}}}&\multicolumn{2}{c}{{\cellcolor{mypink}\textbf{0.008}\stdvu{$\underline{0.013}$}}}\\
\bottomrule
\end{tabular}
\vspace{-15pt}
\end{table*}

\vspace{-4pt}
\subsection{Comparison with State-of-the-arts}\label{sec:acc}
\vspace{-4pt}
% To investigate the effect of our core designs, we conduct comparison experiments with various state-of-the-art FL algorithms under the two scenarios in Table~\ref{tab:1}, \ie, \texttt{In-Federation}, which indicates the testing data follows the same distribution as the training data, and \texttt{Out-of-Federation}, which indicates the testing data is unseen to the local models during training. 

\noindent{\textbf{In-Federation Performance.}}~The first subtable of Table~\ref{tab:1} provides a comprehensive evaluation of the \texttt{In-Federation} setting with regards to various FL algorithms, where \textit{SingleSet} indicates that each client is trained to use their local data without FL, and \textit{Centralized} indicates that all the local datasets are gathered, which serves as the upper-bound of FL models. $\textbf{\texttt{Ours}}$ (\textit{w.}~Pro) and $\textbf{\texttt{Ours}}$ (Full) are the variants of our method that only with prompts and our full method that with both prompts and the null space mechanism. FedAvg (\textit{w.}~FFt)~\cite{mcmahan2017communication} indicates the FedAvg algorithm while employing the pretrained model on the local side and full fine-tuning with all the participating clients. For a fair comparison, we retrain them on the two different scenarios with their default parameters and report their optimal results. All the competing methods are pretrained with Swin Transformers on \textbf{fastMRI} and fully fine-tuned with $50$ communication rounds and $10$ local epochs because they all converge before round $50$. To assess the difference between a pair of methods, we use the paired Student’s $t$-test to demonstrate that all results were at the 0.001 level of statistical significance.

\vspace{-2.5pt}
From Table~\ref{tab:1}, $\textbf{\texttt{Ours}}$ (\textit{w.}~Pro) is superior to the competing FL methods in terms of PSNR, SSIM, and NMSE. In comparison to FedAvg (\textit{w.}~FFt), $\textbf{\texttt{Ours}}$ (\textit{w.}~Pro) increases the PSNR results from $32.14$ dB to $35.29$ dB and the SSIM results from $0.903$ to $0.927$, because the FedAvg (\textit{w.}~FFt) is prone to overfitting when the local data is limited (\ie, issue \myhyperlink{Q1}{\ding{182}}) even though FedAvg (w. FFt) also employ the pretrained model on the local side. However, the method of freezing backbone networks that only fine-tune prompts is consistent with the mechanism of personalized FL, \eg, FedMRI, which divides the local network into a shared global generalized representation and a locally client-specific representation. More importantly, the number of training parameters and communication cost of $~\textbf{\texttt{Ours}}$ (\textit{w.}~Pro) is only $6$\% of FedAvg (\textit{w.}~FFt). 
% Interestingly, $~\textbf{\texttt{Ours}}$ (Full) further improves the reconstruction results (\ie, PSNR: $35.29$ dB$\rightarrow$ $\textbf{36.41}$ dB and SSIM: $0.927$ $\rightarrow$ $\textbf{0.945}$), which is mainly because the federated visual prompts are learned in the null space of global prompts retaining the global knowledge learned in the previous rounds and thereby greatly mitigating catastrophic forgetting (\ie, issue \myhyperlink{Q3}{\ding{184}}). 
Moreover, even compared to FedReg~\cite{xu2022acceleration}, which aims to address catastrophic forgetting in FL, our method still achieves $9.43$\% improvement in terms of PSNR. Furthermore, the reconstruction results of our proposed method are almost on par with the \textit{Centralized} results (upper bound), \ie, PSNR: $36.71$ dB \textit{vs.}~$\textbf{36.41}$ dB, and SSIM: $0.947$ \textit{vs.}~$\textbf{0.945}$.

\begin{figure*}[!t]
    \vspace{-15pt}
	\begin{center}
		\includegraphics[width=0.98\linewidth]{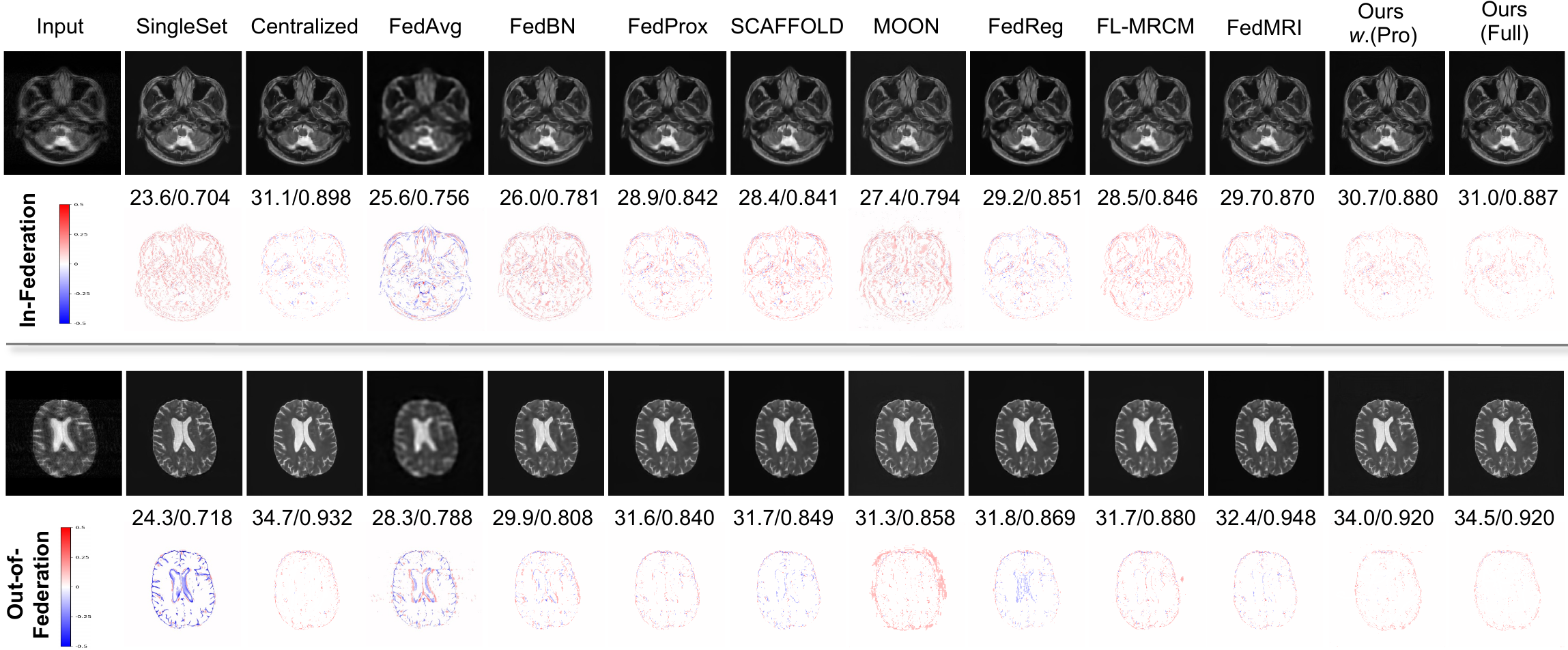}
	\end{center}
    \vspace{-18pt}
	\captionsetup{font=small}
	\caption{\small\textbf{Qualitative comparison} of different algorithms in terms of reconstruction images and error maps with corresponding quantitative measurements in PSNR/SSIM under \texttt{In-Federation} and \texttt{Out-of-Federation} scenarios. The less texture in the error map, the better reconstruction quality.}
	\vspace{-15pt}
	\label{fig:error}
\end{figure*}

In contrast, SingleSet has the lowest results, even though it is also fine-tuned on the pre-trained model, as it cannot solve non-i.i.d. problems. For the classical federated MRI methods, FL-MRCM~\cite{guo2021multi} and FedMRI~\cite{feng2022specificity}, they lead to poor reconstruction performance when the number of local clients is large and the amount of local data is insufficient. Furthermore, other FL algorithms are either designed for data heterogeneity problems or for catastrophic forgetting problems. They have not considered the clinical issues of federated MRI, \ie, issues \myhyperlink{Q1}{\ding{182}}-\myhyperlink{Q3}{\ding{184}}, causing it to produce less accurate reconstructions. These findings confirm our core idea that the three issues of federated MRI can be relieved by learning a federated visual prompt with our federated MRI paradigm. In addition, we visualize the reconstructed images and the corresponding error maps for all the competing methods in the first two rows of Fig.~\ref{fig:error}. The fewer textures in the error map, the better the reconstruction. It is obvious from the error map in Fig.~\ref{fig:error} that our method can significantly reduce the reconstruction error.

\vspace{-2.5pt}
\noindent{\textbf{Out-of-Federation Performance.}}~The second subtable of Table~\ref{tab:1} provides quantitative evaluations under the \texttt{Out-of-Federation} scenario of various FL algorithms. Since the testing distribution in the \texttt{Out-of-Federation} scenario is unseen by the training distribution, the overall results of the \texttt{Out-of-Federation} scenario are slightly lower than those of \texttt{In-Federation}. However, our method still achieves the best performance in terms of the reconstruction results on the three metrics, \ie, $~\textbf{\texttt{Ours}}$ (Full) improves PSNR values from $31.07$ dB to $\textbf{35.60}$ dB, raises SSIM values from $0.899$ to $\textbf{0.939}$, and decreases NMSE values from $0.021$ to $\textbf{0.008}$. Notably, our method even achieves comparable results to the \textit{Centralized} (Upper-bound) method, \ie, PSNR: $35.73$ dB \textit{vs.}~$\textbf{35.60}$ dB, SSIM: $0.940$ \textit{vs.}~$\textbf{0.939}$ and NMSE: $0.008$ \textit{vs.}~$\textbf{0.008}$. That is, our method suffers the least from the issue \myhyperlink{Q3}{\ding{184}} among all competing algorithms. These results support our core idea that learning federated visual prompts in the null space of global prompts can alleviate the three key issues in federated MRI while achieving competitive performance on limited local data. In addition, we show the qualitative evaluation results for the \texttt{Out-of-Federation} scenario in the second rows of Fig.~\ref{fig:error}, which visualizes the high-precision results of our method. As can be seen from this figure, our method provides comparable effects to the \textit{Centralized} one, with clear details of tissue texture and boundaries.

\begin{figure*}[!t]
    \vspace{-6pt}
	\begin{center}
		\includegraphics[width=0.98\linewidth]{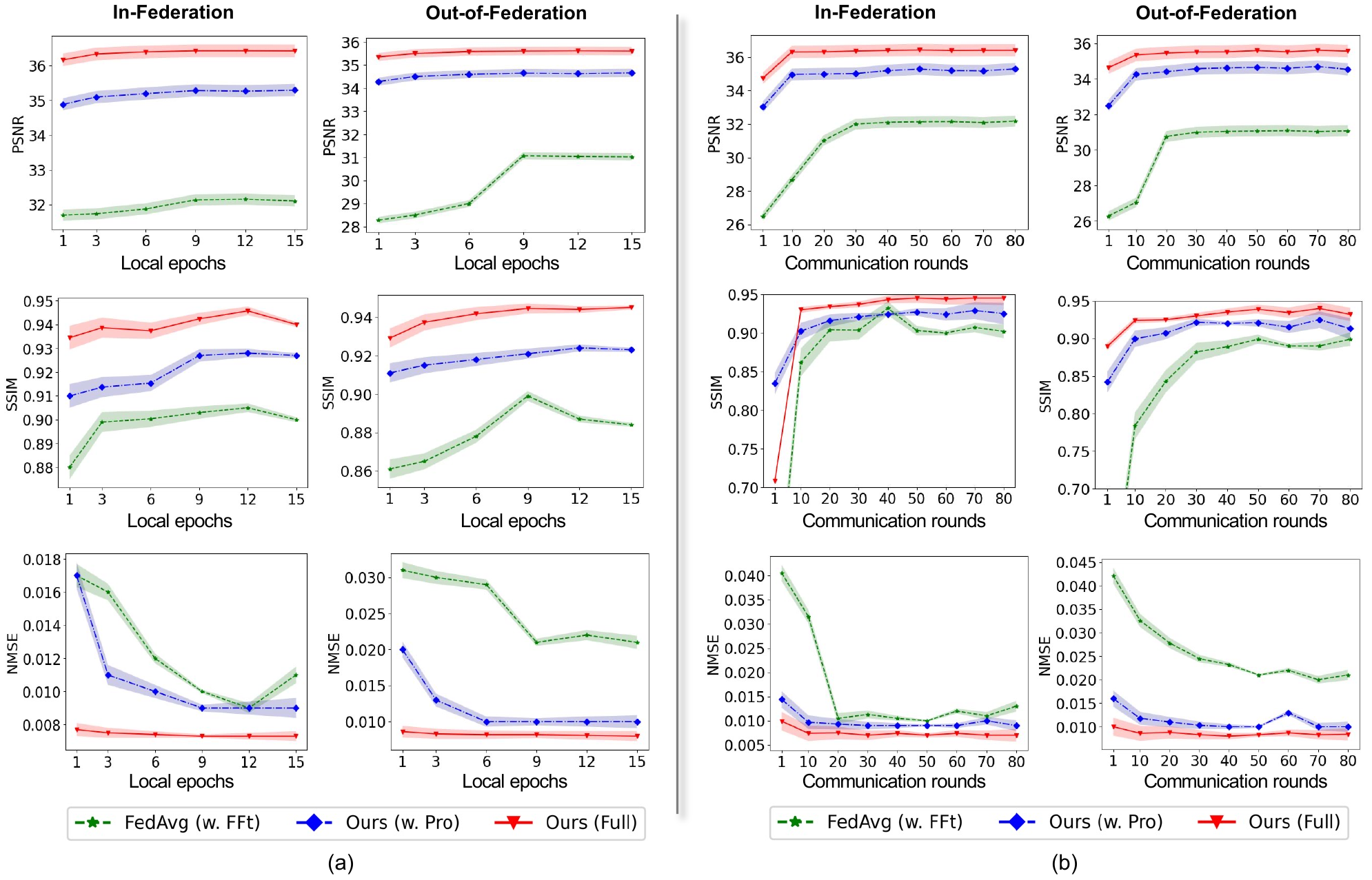}
	\end{center}
    \vspace{-18pt}
	\captionsetup{font=small}
	\caption{\small\textbf{Reconstruction accuracy} of FedAvg (\textit{w.}~FFt), $~\textbf{\texttt{Ours}}$ (\textit{w.}~Pro), and $~\textbf{\texttt{Ours}}$ (Full) versus \textbf{(a)} local epochs and \textbf{(b)} communication rounds under \texttt{In-Federation} and \texttt{Out-of-Federation} scenarios.}
	\vspace{-12pt}
	\label{fig:rou}
\end{figure*}

\begin{figure}[t]
	\vspace{-5pt}
	\begin{center}
		\includegraphics[width=0.95\linewidth]{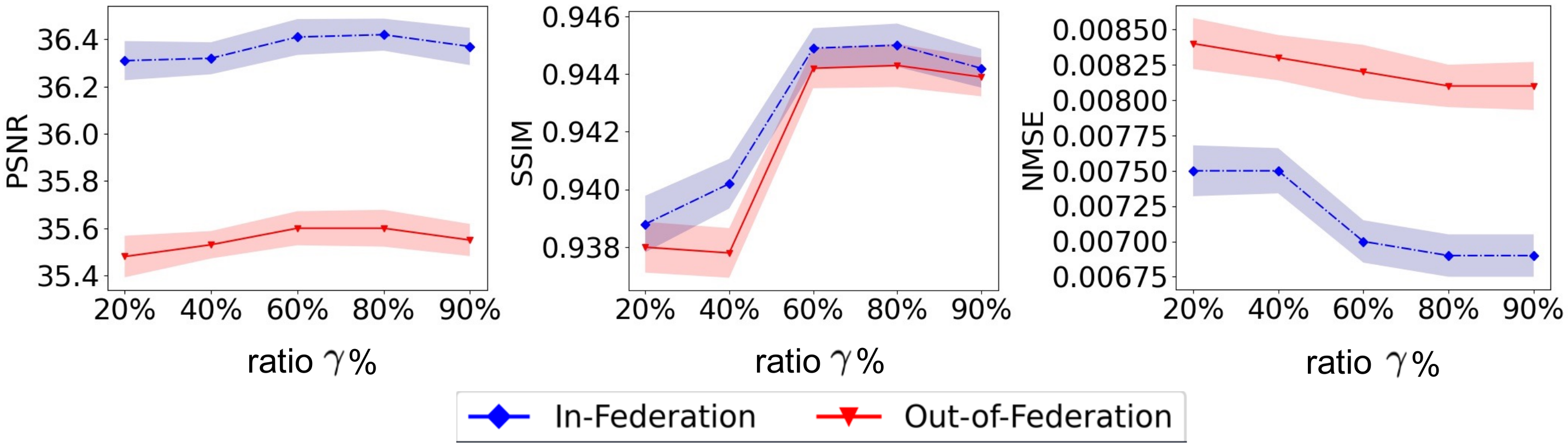}
	\end{center}
	\vspace{-18pt}
	\captionsetup{font=small}
	\caption{\small{Analysis of the \textbf{ratio $\gamma$} \% of the approximate null space in terms of PSNR, SSIM, and NMSE.}}
	\vspace{-22pt}
	\label{fig:ratio}
\end{figure}

\vspace{-4pt}
% \subsection{Knowledge Preservation Analysis} \label{sec:kp}
\subsection{Ablation Studies} \label{sec:kp}
\vspace{-5pt}
\noindent{\textbf{Catastrophic Forgetting \textit{vs.}~Local Updates.}} The catastrophic forgetting (\ie, issue \myhyperlink{Q3}{\ding{184}}) of federated MRI is caused by data heterogeneity due to different imaging protocols. As a result, more local updates will cause the local model to deviate from the global optimal solution. Here, we explore whether more local epochs cause the model to forget previously acquired global knowledge. In other words, it needs to be explored whether our proposed method can force local mitigation of this forgetting. Thus, we record the reconstruction results in terms of PSNR, SSIM, and NMSE for different numbers of local epochs in Fig.~\ref{fig:rou} (a). From this figure, in comparison to FedAvg (\textit{w.}~FFt)~\cite{mcmahan2017communication}, increasing the local epoch does not greatly affect the accuracy of our method, \ie, \textbf{\texttt{Ours}} (\textit{w.}~Pro), and \textbf{\texttt{Ours}} (Full). In particular, as the number of local epochs increases, the performance of our method gradually stabilizes. In contrast, when the number of local epochs is too large, the reconstruction accuracy of FedAvg (\textit{w.}~FFt) will be reduced because it will gradually deviate from the global knowledge area. As a result, the result indicates that our method can learn without forgetting global knowledge over more local epochs.

% \vspace{-3pt}
\noindent{\textbf{Analysis of Ratio $\gamma\%$.}} As we mentioned in Sec.~\ref{sec:null}, the ratio $\gamma\%$ of the smallest singular values in $\boldsymbol{\Lambda}_2^{z+1}$ controls the size of the approximate null space in global prompts. That is, the larger the $\gamma\%$, the smaller the area of knowledge preservation. 
However, the area of knowledge preservation contains the global knowledge from the previous round, which is expected to be unchanged. To analyze the knowledge preservation capability that a good federated MRI framework should have, we discuss the reconstruction accuracy for different $\gamma\%$ values with regard to two scenarios in Fig.~\ref{fig:ratio}. It can be seen from this figure that, with the increase in $\gamma\%$ values, the reconstruction accuracy gradually increases. However, when $\gamma\% = 100\%$, the reconstruction accuracy drops rapidly. This is because when the knowledge preservation area is reduced to zero, fitting on the local distribution leads to forgetting the out-of-local distribution, \ie, the issue \myhyperlink{Q3}{\ding{184}}. Additionally, we can observe from Fig.~\ref{fig:ratio} that roughly $40\%$ of the area comprises previously acquired global knowledge because our method obtains the highest results at this point. Following~\cite{wang2021training}, we use the proportion of the sum of singular values of $\boldsymbol{\Lambda}_2^{z+1}$ in the sum of singular values of $\boldsymbol{\Lambda}^{z+1}$ to verify the rationality of the approximation. We find that the proportion in each layer is smaller than $10^{-7}$. That is, the sum of the smallest singular values $\bm{U}_2^{z+1}$ can be ignored, thereby making it reasonable to approximate the null space through the spatial range of $\bm{U}_2^{z+1}$. Interestingly, our method only performs local updates in the approximate null space of global prompts, which greatly improves the stability of the model and thus prevents catastrophic forgetting, \ie, the issue \myhyperlink{Q3}{\ding{184}}.

\vspace{-2pt}
\noindent{\textbf{Communication Efficiency Analysis.}} Our primal interest is to learn a federated visual prompt that enables FL to perform well with less local data, lower communication cost, and faster convergence. Here, we investigate the communication efficiency of our proposed method in terms of communication cost and different communication rounds, respectively. As shown in Table~\ref{tab:1}, the proposed method only requires $0.11$ M of communication, which is $6\%$ of the others, thereby addressing the issue \myhyperlink{Q2}{\ding{183}}. However, although the classical federated MRI algorithms FL-MRCM~\cite{guo2021multi} and FedMRI~\cite{feng2022specificity} can also achieve good accuracy, they ignored this clinical issue of federated MRI. Additionally, the reconstruction accuracy of FedAvg (\textit{w.}~FFt)~\cite{mcmahan2017communication}$, \textbf{\texttt{Ours}}$ (\textit{w.}~Pro), and $\textbf{\texttt{Ours}}$ (Full) under different communication rounds is recorded in Fig.~\ref{fig:rou} (b), where the number of the local epochs is fixed at $10$ for each method. As can be seen from the figure, our method converges within $10$ rounds, while FedAvg (\textit{w.}~FFt)~\cite{mcmahan2017communication} converges after $30$ rounds. Especially under the mechanism of updating the local prompts in the approximate null space of global prompts, our method provides higher reconstruction results. This is because our federated prompt learning mechanism prevents overfitting caused by the limited amount of local training data and offers a personalized scheme for each client. 
% Moreover, updating local prompts in the null space of global prompts can avoid forgetting previously acquired knowledge and accelerate convergence. 
The results in Fig.~\ref{fig:rou} (b) confirm that our method provides the highest communication efficiency even on limited local data, thus relieving the issues \myhyperlink{Q1}{\ding{182}} and \myhyperlink{Q2}{\ding{183}}.

% \subsection{Federated Prompting Learning}

% 

% \noindent{\textbf{What is the Federated Prompting?}} 

% \noindent{\textbf{How far the local drift from the global model?}}

% \noindent{\textbf{How to balance the global stability and local plasticity?}}

\vspace{-5pt}
\section{Conclusion}
\vspace{-4pt}
%Existing federated MRI algorithms are only designed to deal with data heterogeneity. 
%
% In contrast, this paper propose a new federated paradigm, FedPR, for the three issues of federated MRI, which has the advantages of less communication cost, less local data, and outstanding performance. Benefiting from a powerful pre-trained model, FedPR only learns prompts with a small number of learnable parameters, thus significantly reducing communication costs while achieving competitive performance on limited local data. 
%
This paper presented a federated visual prompt learning method, FedPR, to tackle the three issues in federated MRI reconstruction, \ie, limited communication bandwidth, insufficient local training data, and catastrophic forgetting.
Benefiting from a powerful pre-trained model, our FedPR only learns prompts with a small number of learnable parameters.
Additionally, FedPR greatly mitigates catastrophic forgetting by updating local prompts only in the approximate null space of global prompts, thereby preventing knowledge outside the local distribution from being overwritten.
% and causing catastrophic forgetting. 
Experiments under \texttt{In-Federation} and \texttt{Out-of-Federation} scenarios demonstrate the superiority of FedPR in relieving the three issues of federated MRI reconstruction.

% As discussed, please put the following information into the acknowledgment.
\noindent\textbf{Acknowledgements:}
This work was supported by the Agency for Science, Technology and Research (A*STAR) AME Programmatic Funds (Grant Number: A20H4b0141). This work was also supported by the AI Singapore Tech Challenge (Open-Theme) Funding Scheme (Award No.: AISG2-TC-2021-003).

% MARIO (Grant Number : A20H4b0141): 01/04/2021 – 31/03/2024
% RAPIER (Award No.: AISG2-TC-2021-003): 01 April 2022 to 31 March 2025

%%%%%%%%% REFERENCES
{\small
\bibliographystyle{ieee_fullname}
\bibliography{egbib}
}
\clearpage
\appendix

\setcounter{table}{0}
\renewcommand{\thetable}{A\arabic{table}}
\setcounter{figure}{0}
\renewcommand{\thefigure}{A\arabic{figure}}

% \section*{Contents}
% The following items are included in our supplementary material:
% \begin{itemize}
%   \item Rationality analysis of the approximate null space of the global prompts in Section~\ref{sec:R}.
%   \item Communication efficiency analysis over state-of-the-arts in Section~\ref{sec:com}.
%   \item Additional qualitative results for \texttt{In-Federation} and \texttt{Out-of-Federation} scenarios in Section~\ref{sec:errormap}.
% \end{itemize}

\section{Rationality Analysis of the Approximation}\label{sec:R}

\begin{figure*}[!ht]
\vspace{25pt}
	\begin{center}
		\includegraphics[width=\linewidth]{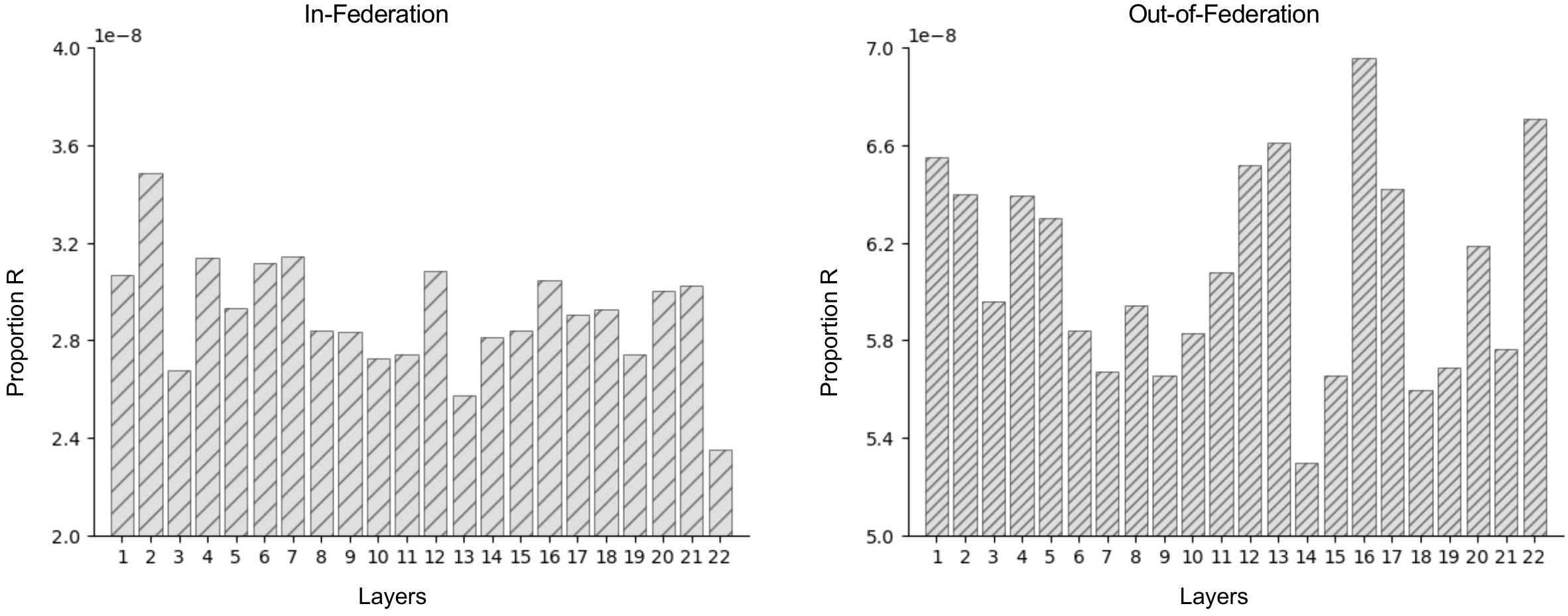}
	\end{center}
% 	\vspace{-16pt}
	\captionsetup{font=small}
	\caption{\small\textbf{Rationality analysis} of the approximate null space under \texttt{In-Federation} and \texttt{Out-of-Federation} scenarios.}
	\label{fig:apR}
\end{figure*}

To verify the rationality of the approximate null space of the global prompts~\cite{wang2021training}, we denote the proportion of the sum of singular values of $\boldsymbol{\Lambda}_2^{z+1}$ in the sum of singular values of $\boldsymbol{\Lambda}^{z+1}$ as
\begin{equation}
R=\frac{\sum{\operatorname{diag}\left\{\boldsymbol{\Lambda}_2^{z+1}\right\}} }{\sum{ \operatorname{diag}\left\{\boldsymbol{\Lambda}^{z+1}\right\}} },
\end{equation}
where ``$\operatorname{diag}$'' indicates the diagonal elements. If the value of $R$ is very small, the sum of the smallest singular values $\bm{U}_2^{z+1}$ can be ignored, allowing the null space to be approximated through the spatial range of $\bm{U}_2^{z+1}$. We record the $R$ values of different layers under the \texttt{In-Federation} and \texttt{Out-of-Federation} scenarios in Fig.~\ref{fig:apR}. As can be seen from this figure, the proportion $R$ in each layer is smaller than $10^{-7}$, indicating that the selected $\bm{U}_2^{z+1}$, which is last $\gamma\%$ of singular values in $\boldsymbol{\Lambda}_2^{z+1}$, can be ignored. Additionally, we observe that the value of $R$ in the \texttt{Out-of-Federation} scenario is  slightly greater than the value of $R$ in the \texttt{In-Federation} scenario. We infer that this is primarily due to the fact that the distribution of the test set in the \texttt{Out-of-Federation} scenario is never seen by the training set, resulting in more severe catastrophic forgetting and thus a slightly higher $R$ value. However, the values of $R$ are still quite close to $0$ in both two scenarios. As a result, our selected $\bm{U}_2^{z+1}$ is a reasonable approximation of the null space of the global prompt.

\section{Communication Efficiency Analysis over SOTAs}\label{sec:com}

\begin{figure*}[!t]
    \vspace{26pt}
	\begin{center}
		\includegraphics[width=\linewidth]{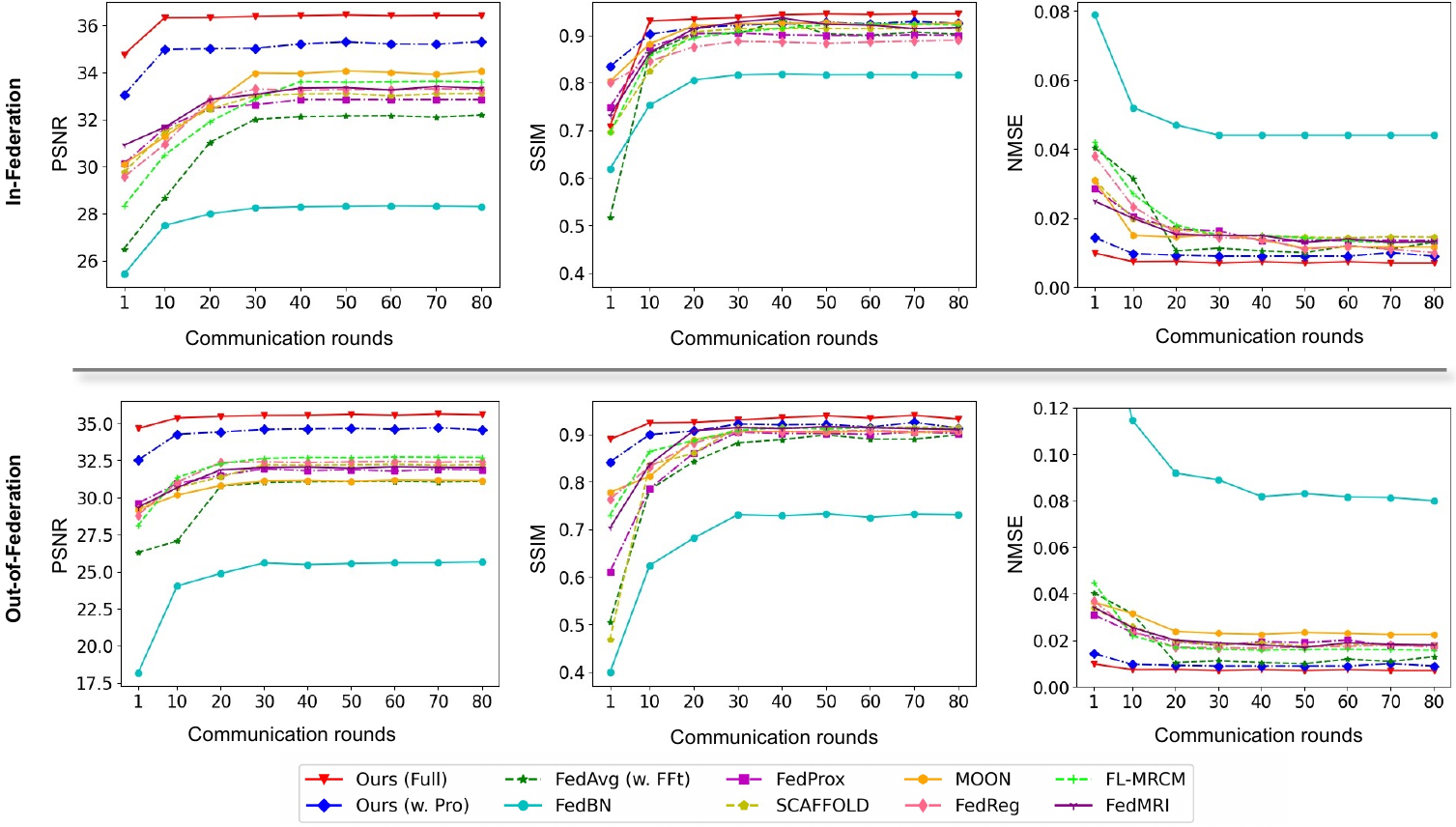}
	\end{center}
    % \vspace{-13pt}
	\captionsetup{font=small}
	\caption{\small\textbf{Communication Efficiency Analysis} over state-of-the-arts versus PSNR, SSIM, and NMSE under \texttt{In-Federation} and \texttt{Out-of-Federation} scenarios.}
% 	\vspace{15pt}
	\label{fig:apcom}
\end{figure*}

To further evaluate our federated visual prompt mechanism regards to tackle the issue \hypertarget{Q3}{\ding{184}} \textit{catastrophic forgetting}, we record the reconstruction accuracy over state-of-the-arts under different communication rounds in Fig.~\ref{fig:apcom}. The number of the local epochs of each method is fixed at $10$. As can be seen from this figure, all the baseline algorithms reached stability after $30$ rounds while our method converges within $10$ rounds. Although these methods also adopt the same pre-trained model as ours, the catastrophic forgetting due to the data heterogeneity mechanism still leads to slower convergence. For example, FedBN~\cite{li2021fedbn} (see the line \protect\includegraphics[scale=0.20,valign=c]{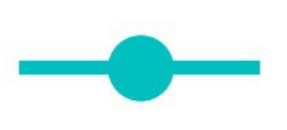} in Fig.~\ref{fig:apcom}) applies the batch normalization on each local client to alleviate the client-shift. However, when there are only a few local training data, batch normalization is performed at the feature level, still resulting in deviations and affecting convergence. Especially in the \texttt{Out-of-Federation} scenario, where testing data is unseen by the local models during training, the performance degrades significantly. However, the convergence speed for the FedReg~\cite{xu2022acceleration} (see the line \protect\includegraphics[scale=0.11,valign=c]{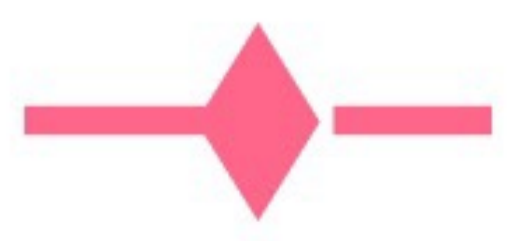} in Fig.~\ref{fig:apcom}), which tries to minimize catastrophic forgetting, is still subpar because the full fine-tuning mechanism still introduces some deviation. On the other hand, FedReg~\cite{xu2022acceleration} adds the regularization loss on the local side increasing the complexity of the algorithm. In contrast, our method updates the local prompts in the approximate null space of global prompts, which can avoid forgetting previously acquired knowledge and accelerating convergence (see the line \protect\includegraphics[scale=0.43,valign=c]{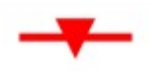} in Fig.~\ref{fig:apcom}).

% \myhyperlink{Q1}{\ding{182}}-\myhyperlink{Q3}{\ding{184}}

% \hypertarget{Q1}{\ding{182}} \textit{insufficient amount of local training data}, 

% and \hypertarget{Q2}{\ding{183}} \textit{limited communication bandwidth}, 

\begin{figure*}[!ht]
% \vspace{-10pt}
	\begin{center}
		\includegraphics[width=0.91\linewidth]{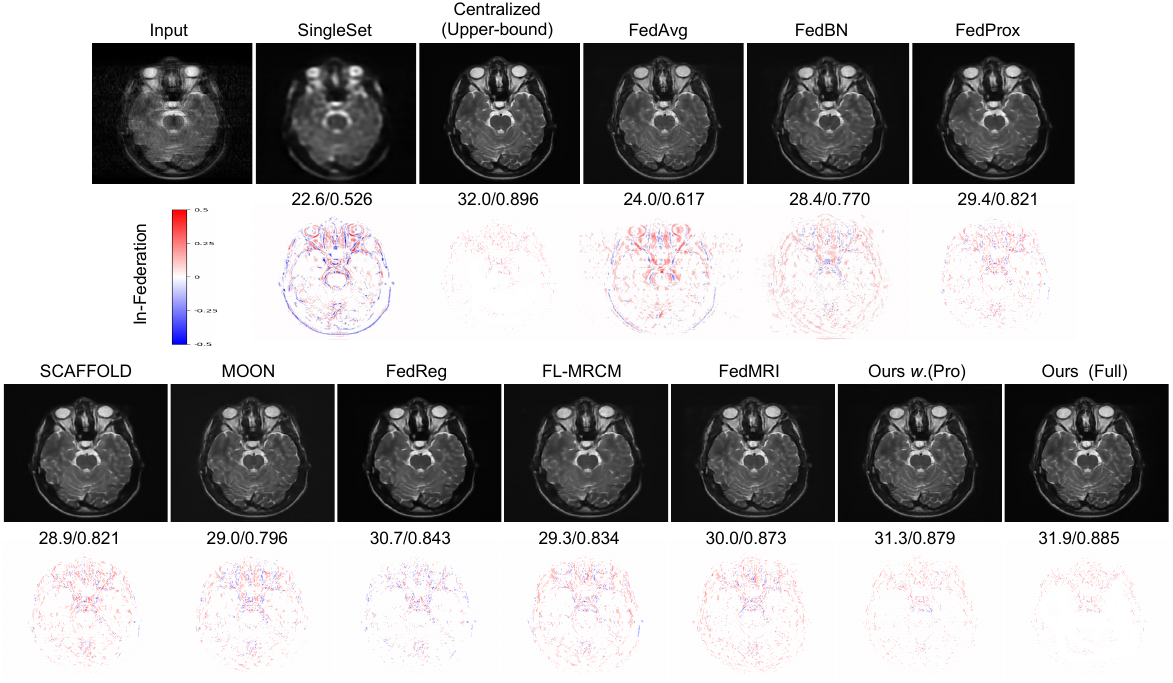}
	\end{center}
	\vspace{-10pt}
	\captionsetup{font=small}
	\caption{\small\textbf{Qualitative comparison} of different algorithms in terms of reconstruction images and error maps with corresponding quantitative measurements in PSNR/SSIM under \texttt{In-Federation} scenario. The less texture in the error map, the better reconstruction quality.}
	\label{fig:aperror1}
\end{figure*}

\begin{figure*}[!ht]
\vspace{8pt}
	\begin{center}
		\includegraphics[width=0.91\linewidth]{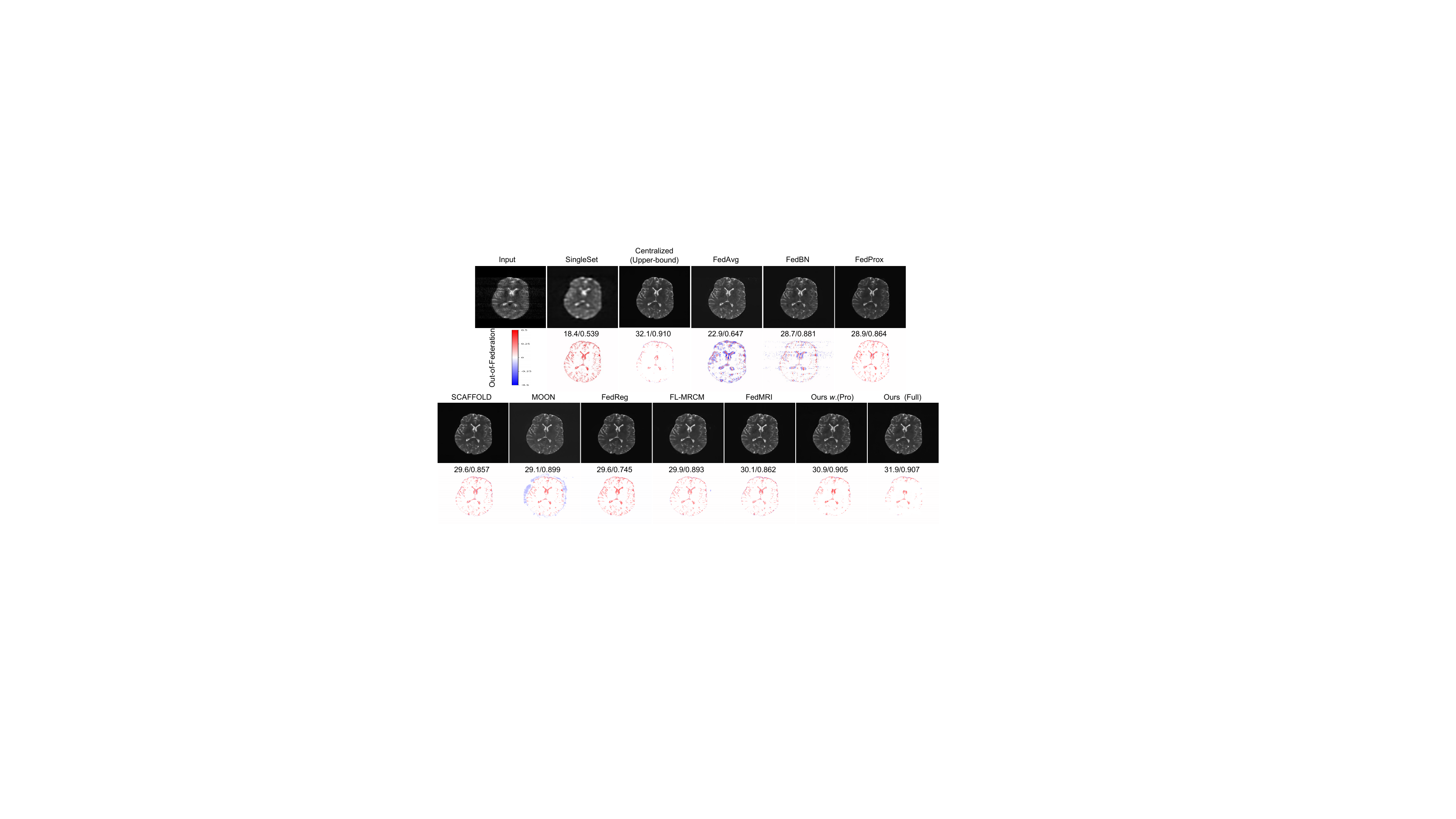}
	\end{center}
	\vspace{-10pt}
	\captionsetup{font=small}
	\caption{\small\textbf{Qualitative comparison} of different algorithms in terms of reconstruction images and error maps with corresponding quantitative measurements in PSNR/SSIM under \texttt{Out-of-Federation} scenario. The less texture in the error map, the better reconstruction quality.}
	\label{fig:aperror2}
\end{figure*}

\section{Additional Qualitative Results}\label{sec:errormap}
We provide additional qualitative improvements with regard to the reconstructed images with their PSNR and SSIM values and corresponding error maps in Fig.~\ref{fig:aperror1} and Fig.~\ref{fig:aperror2}. 
The results are consistent with our previous results, \ie, our method provides the best-quality reconstructed images, and error maps with the least texture, indicating that our FedPR is effective in relieving the issues \hypertarget{Q1}{\ding{182}}-\hypertarget{Q3}{\ding{184}}.

% {\small
% \bibliographystyle{ieee_fullname}
% \bibliography{egbib}
% }

\end{document}

\end{document}